\documentclass{article} 

\usepackage{iclr2016_conference,times}
\usepackage{hyperref}
\usepackage{url}

\usepackage{mathptmx} 

\newcommand{\ignore}[1]{}
\usepackage{fancyhdr}
\usepackage[normalem]{ulem}
\usepackage{hyperref}
\usepackage{color}
\usepackage{caption}
\usepackage{subcaption}
\usepackage{amsmath}
\usepackage{multirow}
\usepackage{graphicx}
\usepackage{float}
\usepackage{xcolor}
\usepackage{xspace}

\title{Reduced-precision strategies for Bounded Memory in Deep Neural Nets} 
\author{
Patrick Judd$^1$, Jorge Albericio$^1$, Tayler Hetherington$^2$, Tor Aamodt$^2$,  \\
{\bf Natalie Enright Jerger$^1$, Raquel Urtasun, $^3$ and Andreas Moshovos$^1$}  \\
\\
$^1$Department of Electrical and Computer Engineering \\
University of Toronto \\
\texttt{\{juddpatr,jorge,enright,moshovos\}@ece.utoronto.ca} \\
\\
$^2$Department of Electrical and Computer Engineering \\
University of British Columbia \\
\texttt{\{taylerh,aamodt\}@ece.ubc.ca}
\\
\\
$^3$Department of Computer Science \\
University of Toronto \\
\texttt{urtasun@cs.utoronto.edu}
}

\begin{document}
\maketitle

\begin{abstract}

This work investigates how using reduced precision data in Convolutional Neural Networks (CNNs) affects network accuracy during classification.
More specifically, this study considers networks where each layer may use different precision data.
Our key result is the observation that the tolerance of CNNs to reduced precision data not only varies \textit{across} networks, a well established observation, but also \textit{within} networks.
Tuning precision per layer is appealing as it could enable energy and performance improvements.
In this paper we study how error tolerance across layers varies and propose a method for finding a low precision configuration for a network while maintaining high accuracy. 
A diverse set of CNNs is analyzed showing that compared to a conventional implementation using a 32-bit floating-point representation for all layers, 
and with less than 1\% loss in relative accuracy, the data footprint required by these networks can be reduced by an average of 74\% and up to 92\%.

\end{abstract}

\section{Introduction}

Deep learning approaches attempt to learn complex high level abstractions of the data by composing simple non-linear transformations.  
Convolutional Neural Networks (CNNs) have been shown to be extremely effective when solving supervised learning tasks, particularly in the context of object recognition, e.g., detection \citep{girshick2013rich,overfeat}, segmentation \citep{ChenARXIV2015b}, image classification \citep{cudaconvnet}. 

Current approaches utilize very deep architectures, which require massive amounts of memory.
As a consequence it is difficult to train networks without exploiting strategies such as sample or model parallelism.
More importantly, in order to make these networks applicable to real-time processing in small devices such as embedded systems or mobile devices, the memory requirements have to be addressed. 

A popular approach has been to {\it reduce the precision of the data}, that is the number of bits used.
While this introduces approximation error in the internal calculations, CNNs have proven to be very tolerant \citep{arc}.
This is probably due to the fact that training is robust to noise, e.g., by  using mini batches, data augmentation, and/or dropout \citep{dropout}.
Software implementations of CNNs, including GPU accelerated ones, commonly use single-precision floating point values \citep{caffe, GCN, PASCAL}.
However, it is recognized that the large dynamic range of floating point values in unnecessary, and that a fixed point representation might suffice.
The same is true for the 32-bit precision, as most implementations use only 16 bits \citep{DaDiannao,GuptaAGN15,CourbariauxBD14}. 
Further reducing the precision has also been explored \citep{kyukeon_fixed_2014,anwar_fixed_2015,binaryconnect,nnfm}.

Using a reduced precision representation has many benefits: 
1) saves energy in the memory and communication channels (e.g., memory and on-chip links), 
2) improves performance in memory bound systems through better memory bandwidth utilization and effective cache capacity, and 
3) supports larger networks on systems with a fixed memory budget. 

Previous work has primarily used a one-size-fits-all approach to choosing precision, resulting in a precision length that is short enough to work well for \textit{all} networks. 
This is a \textit{worst case} analysis approach where all networks are forced to use the longest representation needed by \textit{any} network among those under consideration.

In contrast, this work analyzes the tolerance of CNNs to reduced precision error at a \textit{per-layer} granularity. It first corroborates that there is significant variance in the precision needed for the weights \textit{across} different networks and layers. This observation is the first contribution of our work. 
Building upon this observation, our second contribution is a method for selecting per layer precisions that maintain network accuracy within a desired range.

We study the effects of per-layer precision selection in the context of five well known CNN architectures: LeNet \citep{lenet}, Convnet \citep{cudaconvnet}, AlexNet \citep{AlexNIPS2012}, Network in Network (NiN) \citep{nin}, and GoogLeNet \citep{goingdeeper14}. 
We show that to maintain accuracy within 1\% of the full precision obtained using a network with 32-bit single-precision floating-point values, layers can instead use a fixed-point representation of only 14 bits in the worst case and of just 2 bits in the best case.
We demonstrate that allowing each layer to use a different precision can reduce the memory footprint of a network on average by 76\% when compared to using 32-bit values, or by 51\% when compared to 16-bit values. These results serve as motivation for pursuing further work in implementing such memory and computation optimizations.

The rest of this paper is organized as follows.
Section \ref{sec:motivation} reports an analysis of the per layer error tolerances of five networks, follow by  a method for finding the best mixed representation for a given network. %
Section \ref{sec:relatedwork} discusses related work, and  
Section \ref{sec:conclusion} summarizes our observations and results.


%

\section{CNN Accuracy vs. Representation Length}
\label{sec:motivation}
\newcommand{\repfigscale}{0.30}
This section studies the data representation length requirements for our targeted CNNs.
Section~\ref{sec:analysis-metho} details the experimental setup and measurement methodology, defines the terminology used throughout the rest of this study, and lists the CNNs studied. Section~\ref{sec:worstcaseanalysis} investigates how accuracy varies with precision across networks where all layers within each network are forced to use the same representation. This analysis corroborates that precision requirements vary across networks. Section~\ref{sec:perlayerneeds} studies per layer precision requirements for each layer in isolation demonstrating that the precision needs vary within each network, the key result of this study. Here we study each layer in isolation varying precision one layer at a time. Finally, 
Sections~\ref{sec:datatraffic} and~\ref{sec:optimal} consider the effects of per layer precision selection to overall network accuracy where we may assign a different precision to each layer. This study is done in the context of memory traffic reduction. Accordingly, Section~\ref{sec:datatraffic} reports the baseline memory traffic requirements whereas Section~\ref{sec:optimal} proposes a method for selecting per layer precisions while maintaining overall network accuracy within a desired range of that of the baseline. %

\subsection{Measurement Methodology}
\label{sec:analysis-metho}

\textbf{CNN Library:} The results of this section are collected using the popular Caffe framework \citep{caffe}. To measure the effects of reduced precision we used source code level modifications, a method that precluded using closed-source implementations. However, the conclusions drawn about precision variation tolerance should be applicable to other implementations of CNNs that use a 32-bit floating-point representation. The different CNNs are implemented in Caffe using a 32-bit single-precision floating-point representation for all numerical data.

\textbf{How was Precision Varied per Layer:} To study the effect of numerical representation on accuracy, we convert the values to the desired representation  and then back to single-precision floating-point prior to processing them in each layer.
This is appropriate for any potential memory and communication optimizations that would  not change the way computations are performed but rather how data is represented when communicated across layers either on-chip or through memory.

To perform this analysis we modified the Caffe framework to capture data read and write calls and convert the default single-precision floating-point values into a lower precision fixed-point representation.
Note that precision is lost during this conversion.
Prior to starting the computation for each layer, we convert the fixed-point numbers into single-precision floating point.
This conversion does not restore the original floating point number, as precision was lost during the first conversion.

\textbf{Target Numerical Representation:} We target a fixed-point representation for all values processed by the networks and study the length required for accurate classification.
Fixed-point representations are compatible with integer arithmetic units and conversion from existing numerical data types is relatively straightforward.  We parameterize an $N$-bit fixed point value as having  $I$ integer bits and $F$ fractional bits. We study how changing $I$ and $F$ across layers and networks affects the overall network accuracy. 

\textbf{Values Studied:} We consider both the model values and the layer data outputs/inputs to each layer during classification. We will use the terms \textbf{weights} to refer to the model weights (after training) and \textbf{data} to refer to the  output of each layer.

\begin{table*}[t]
\centering
\begin{tabular}{|l|l|l|l|r|r|}
\hline
\textbf{Task} & \textbf{Data set} & \textbf{Network} & \textbf{Layer} & \textbf{Top-1 Accuracy} \\ \hline
Digit classification &MNIST & LeNet &  2 CONV and 2 FC & 0.9904 \\
\hline
Image classification &CIFAR10  & Convnet & 3 CONV and 2 FC & 0.7173 \\
\hline
\multirow{3}{*}{Image classification}&\multirow{3}{*}{ImageNet}  & AlexNet &  5 CONV and 3 FC & 0.5748 \\ \cline{3-5}
&& NiN & 12 CONV & 0.5592 \\ \cline{3-5}
&& GoogLeNet & 2 CONV and 9 IM & 0.6846 \\ \hline
\end{tabular}
\caption{Networks studied: Accuracy reported is for the baseline configuration. CONV = convolution, FC = fully connected, IM = inception module. 
Appendix \ref{sec:appendix} describes the layers in more details.}
\label{table:networks}
\end{table*}

\textbf{CNNs Studied:} We consider the five most popular neural networks used for image classification which are listed in Table~\ref{table:networks} along with their respective datasets. 
They range from the relatively simple five layer LeNet to the 22 layer GoogLeNet which was the best network in the 2014 ImageNet Competition \citep{ILSVRC15}. 

We use the models and the pre-trained weights that are available for these networks either through the Caffe distribution or the Caffe Model Zoo \citep{model-zoo}. 
For ImageNet we use the ILSVRC2012 Task 1 dataset \footnote{http://www.image-net.org/challenges/LSVRC/2012/}.
We run each network for 100 batches of 50-100 input images from the validation set of the respective dataset. 
The last column of Table~~\ref{table:networks} reports the accuracy achieved on the baseline Caffe implementation, which uses single-precision floating-point values.

\textbf{Accuracy Metric:} We use top-1 accuracy (instead of the typical top-5) to increase the sensitivity to reduced precision error. 
Further reduction in precision may be possible if we were to consider the top-5 accuracy. Table~\ref{table:networks} reports the baseline top-1 accuracy for the CNNs considered.

\begin{figure}[t]
\centering
\includegraphics[width=0.3\textwidth]{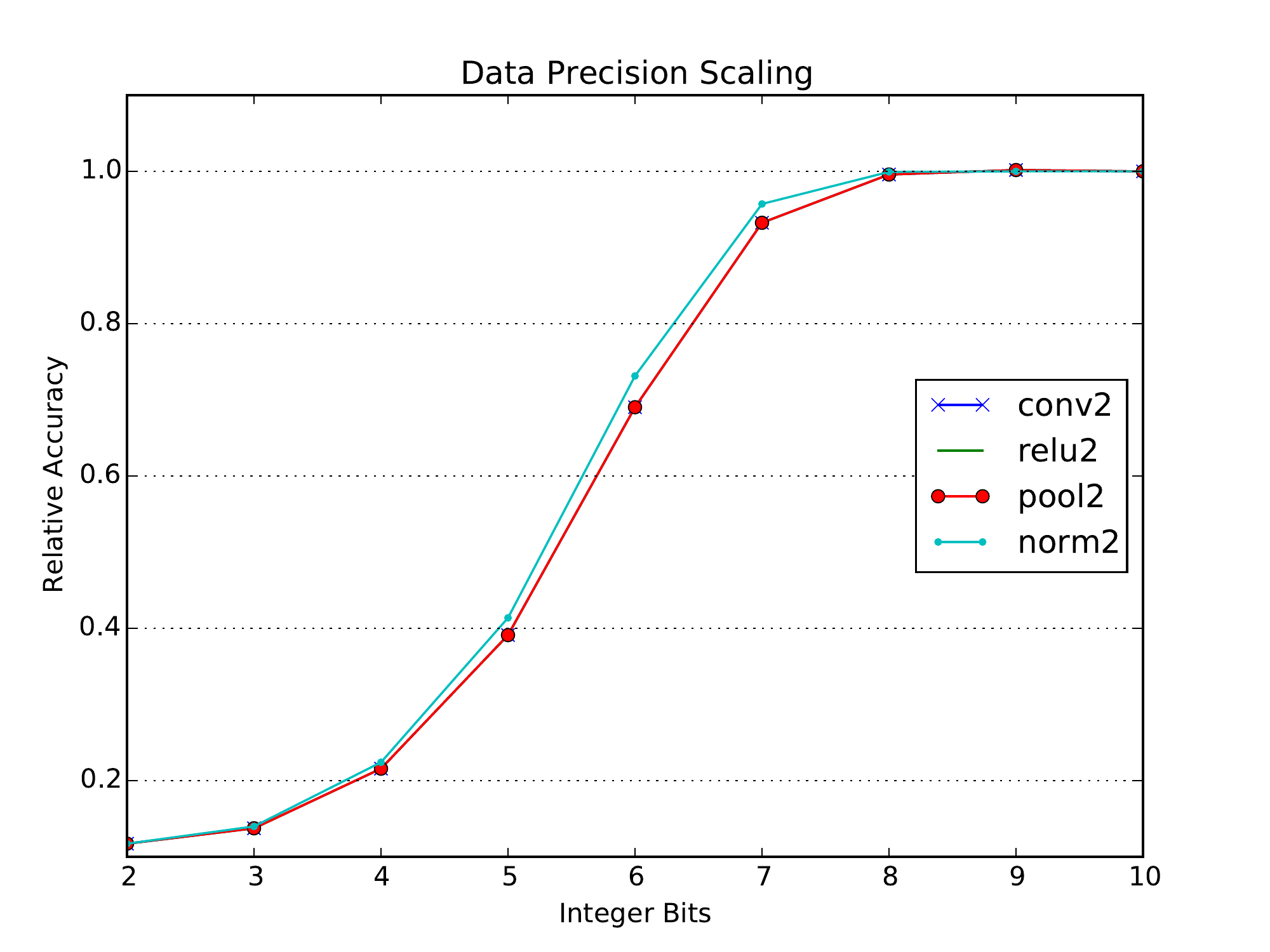}
\caption{AlexNet accuracy variation as a function of data bits within the second convolution layer.}
\label{fig:alexnet-conv2}
\end{figure}

\textbf{Assigning Precision:} For most networks we consider assigning a particular precision to each layer. 
We could go a step further and assign a precision to each computational stage within the layer, however, we have found stages within the same layer tend to have the same precision tolerance. 
In support of this observation, Fig. \ref{fig:alexnet-conv2} shows evidence that the precision requirements within the individual computational stages 
of the second convolution layer of  AlexNet has very similar precision requirements.  
For GoogLeNet, we assign a precision to each "inception module" to simplify the analysis and refer to them as "layers" to be consistent with the other networks.

A layer typically contains a single, convolution or fully-connected stage which performs the bulk of the computation.
This stage is often followed by a series of simpler stages such as ReLU, pooling and linear response normalization.
Activation functions like ReLU are often considered part of the previous stage but for consistency we follow Caffe's model where activations are a separate stage.

Table~\ref{table:networks} reports the  number and type of layers per network. There are three layer types according to the main layer that affects accuracy: 
1) CONV where the main stage is a convolution, 2) FC where the main stage is a fully-connected layer, and 3) IM, for GoogLeNet, where we treat inception modules as layers. 
Appendix \ref{sec:appendix} lists the stages in each layer of each network.

\subsection{Uniform Representation Across All Layers}
\label{sec:worstcaseanalysis}
The results of this section confirm that precision requirements vary \textit{across} networks. 
Specifically, this section studies the per network, minimum \textit{uniform} representation length.
For this analysis, we require that the same representation be used by all layers in the network.
Since existing implementations choose a presentation that is sufficient for \textit{any} network, they use a \textit{worst case} analysis approach.
The results of this section demonstrate, that this current \textit{worst case} analysis is suboptimal.

\begin{figure*}[t!]
\centering
\begin{subfigure}{.3\linewidth}
{\includegraphics[width=\textwidth]{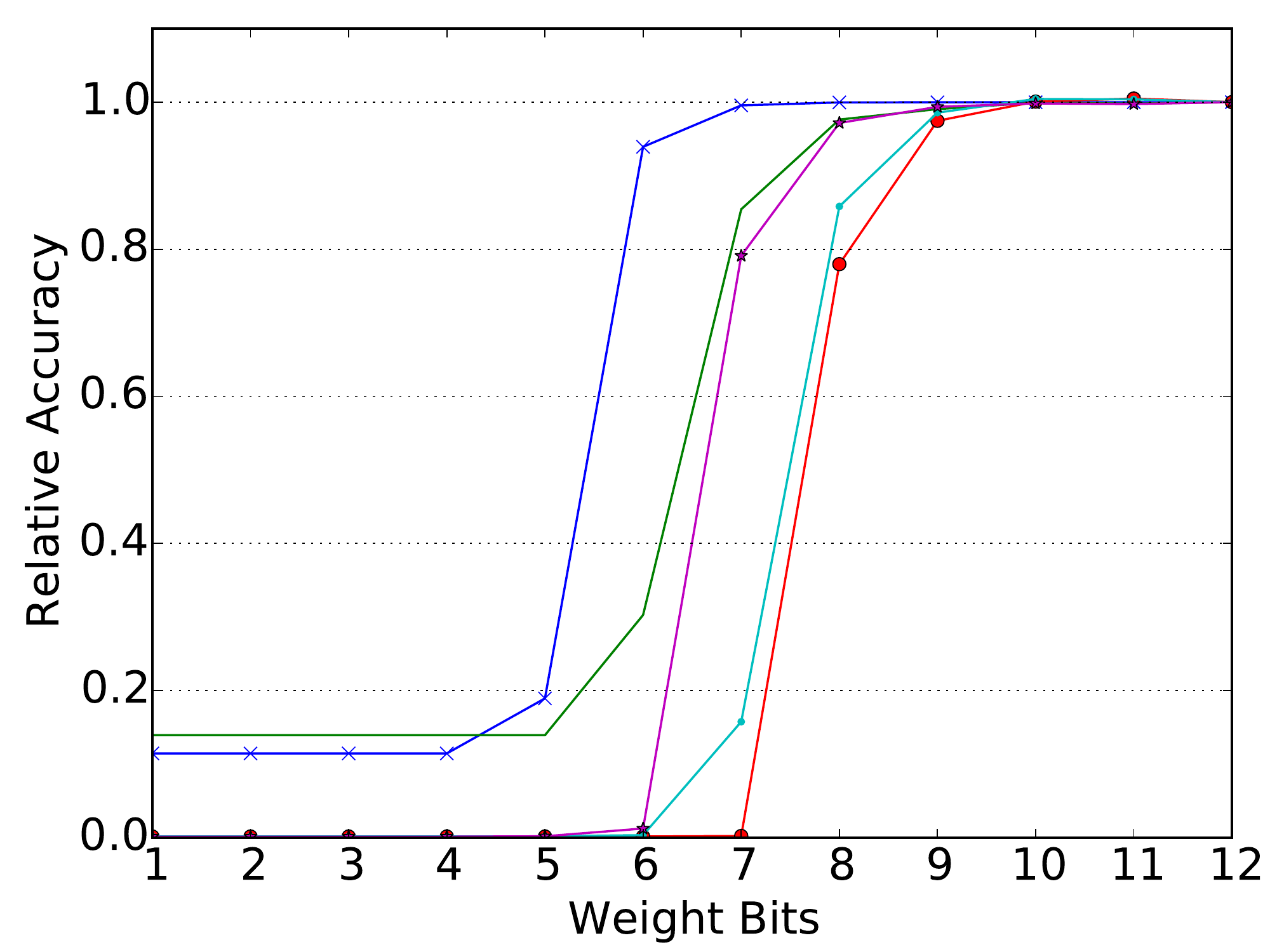}}%
\caption{Weights}
\end{subfigure}
\begin{subfigure}{.3\linewidth}
{\includegraphics[width=\textwidth]{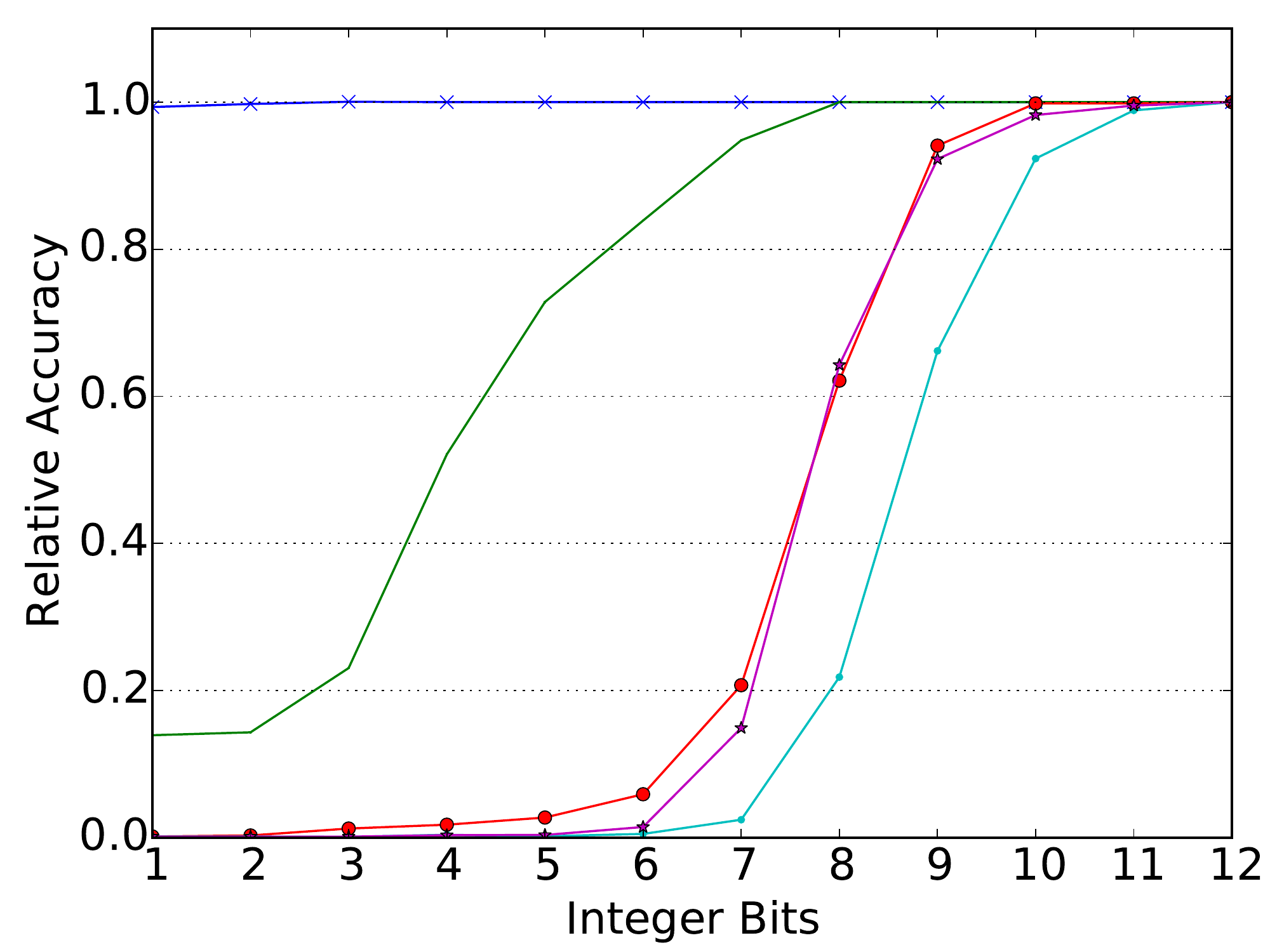}}%
\caption{Data: Integer}
\end{subfigure}
\begin{subfigure}{.3\linewidth}
{\includegraphics[width=\textwidth]{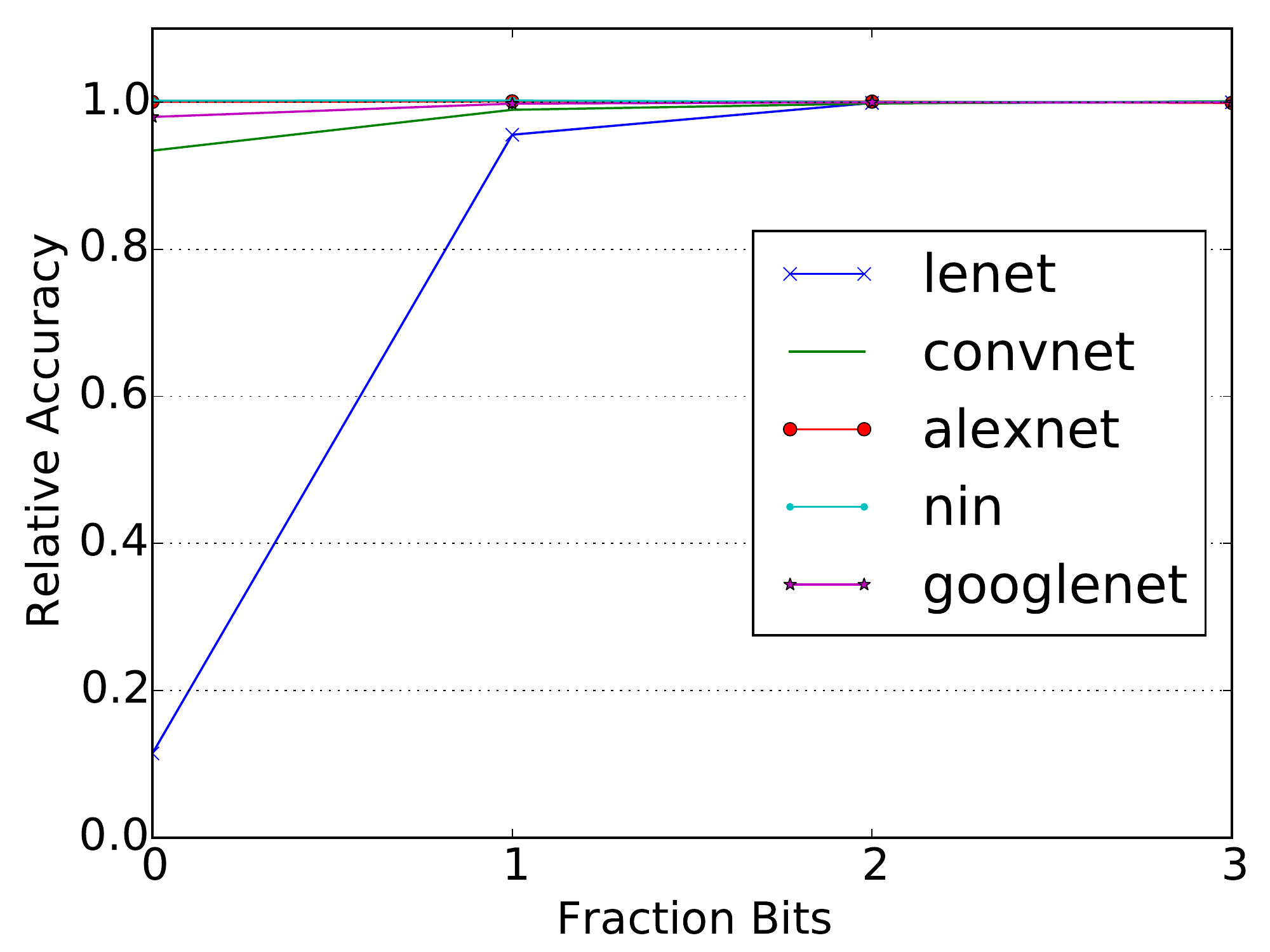}} 
\caption{Data: Fraction}
\end{subfigure}
\caption{
Accuracy relative to the baseline when the bit width is uniform for all layers.
}
\label{fig:4nets-uniform}
\end{figure*}

\noindent{\bf Weights:} Fig. ~\ref{fig:4nets-uniform}(a) shows accuracy variation with the number of bits used for the weights.
Weights real numbers, typically between -1 and 1, and hence we fix the integer part to 1 bit and only report results when varying the fractional part of the fixed-point representation.
Using 10-bit weights is sufficient to maintain accuracy for all networks studied.
Note that LeNet can use 8-bit fixed-point weights as well with no loss in accuracy.

\noindent{\bf Data:} Figs ~\ref{fig:4nets-uniform}(b) and (c) report accuracy  as we vary the number of bits used for the integer (b) and fractional (c)  portions of the fixed-point representation for the data.
Accuracy in Convnet and LeNet persists when the integer portion is at least 8 bits, whereas the other networks require at least 11 bits.
Fig.~\ref{fig:4nets-uniform}(c) shows that most networks need just one fractional bit and some require at most two.
These results suggest that a uniform fixed-point representation for the intermediate data values flowing through the network will require a 14-bit fixed-point representation, with 12 integer and 2 fractional bits.

If we had to chose a uniform fixed-point representation for the weights and the data, 
it would have to be at least 12 bits for the integer portion to accommodate the needs of the intermediate data values, 
whereas the fractional portion would have to be at least 9 bits to accommodate the fractional portion of the weights.
In total we would need 21 bits.
This result corroborates the findings of past work that suggested using fixed-point representations \citep{155324,324283}.
The results of this section also support the design choices of recent accelerators that use a 16-bit fixed-point representation with little loss in accuracy \citep{diannao, DaDiannao}.

\begin{figure*}[t]
\centering
\begin{subfigure}{.3\linewidth}
\def \mynet {lenet}
{\includegraphics[trim=0cm 0.75cm 0cm 1.1cm, clip=yes, height=0.6\linewidth]{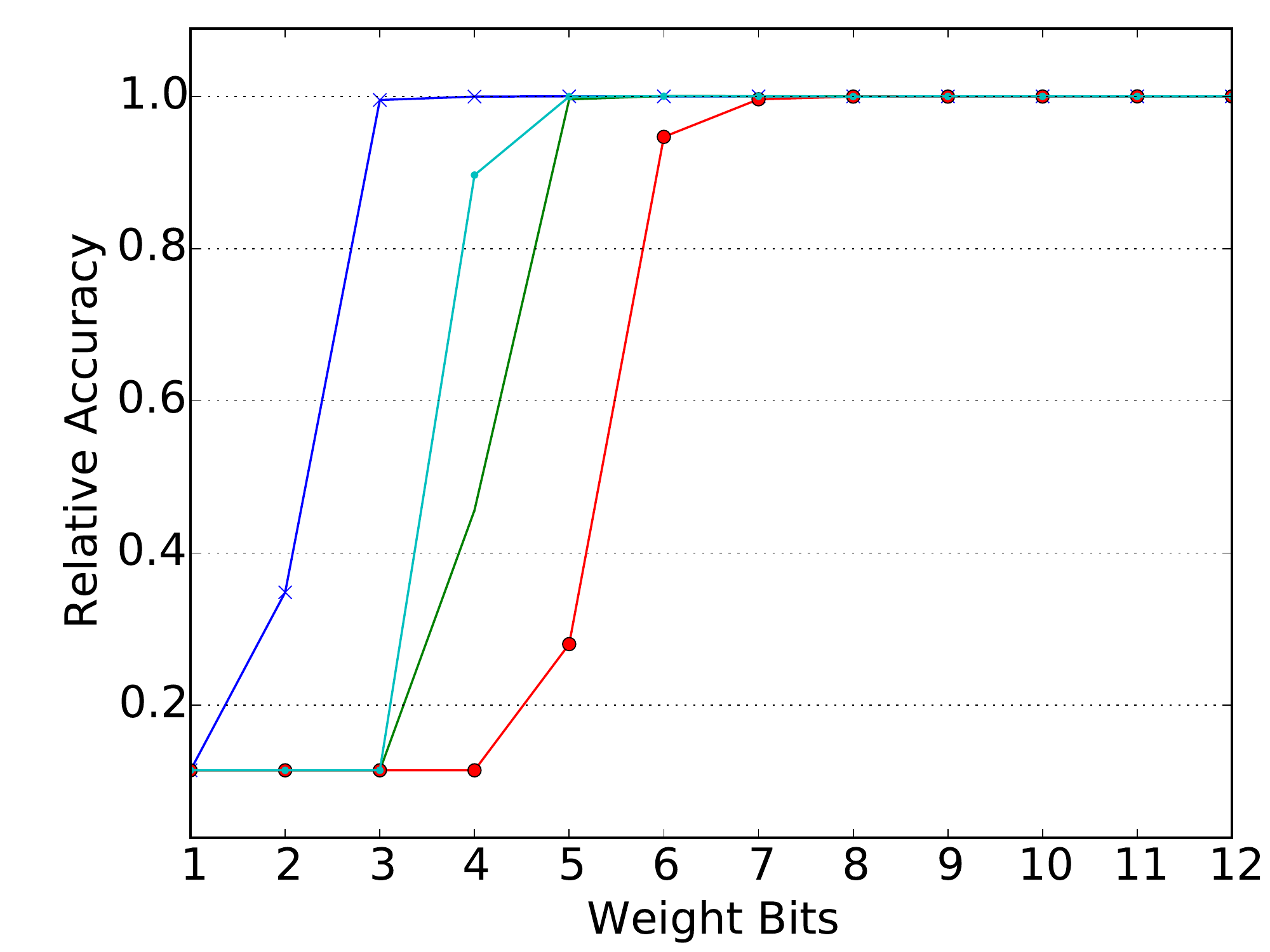}} 
\caption{LeNet: Weights}
\end{subfigure}
\begin{subfigure}{.3\linewidth}
\def \mynet {lenet}
{\includegraphics[trim=0cm 0.75cm 0cm 1.1cm, clip=yes, height=0.6\linewidth]{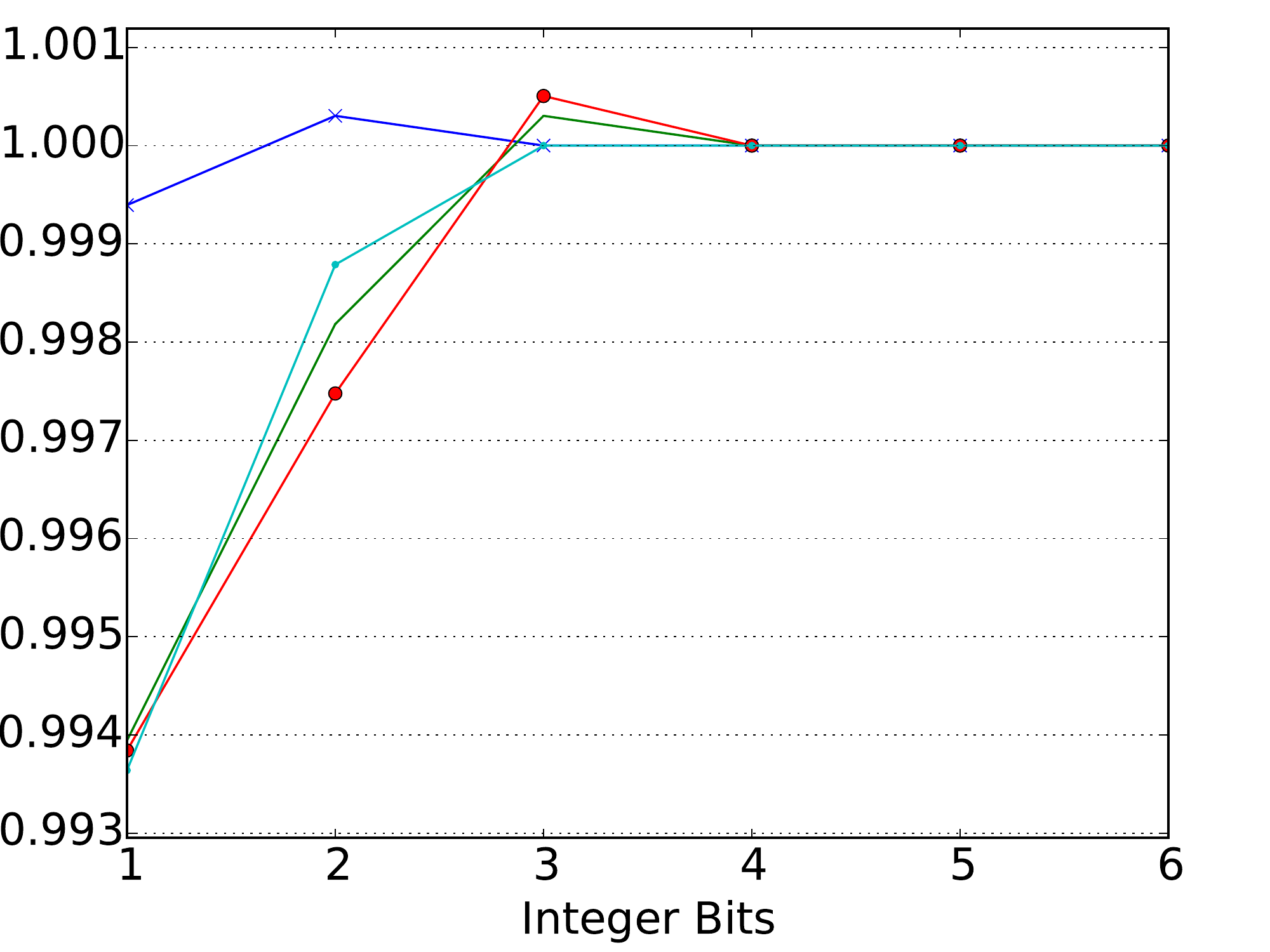}}%
\caption{LeNet: Data: Integer}
\end{subfigure}
\begin{subfigure}{.3\linewidth}
\def \mynet {lenet}
{\includegraphics[trim=0cm 0.75cm 0cm 1.1cm, clip=yes, height=0.6\linewidth]{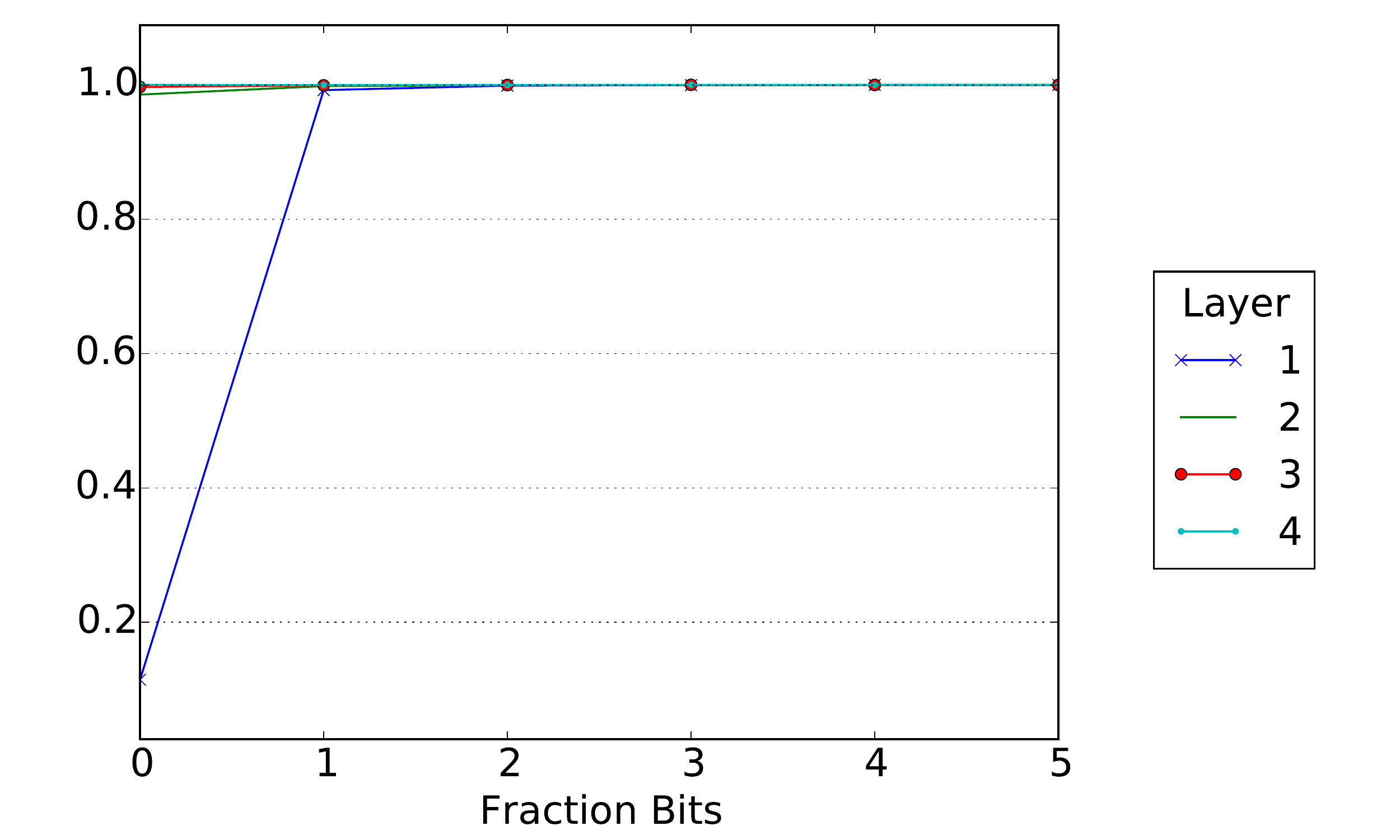}}
\caption{LeNet: Data: Fraction}
\label{fig:perlayerlenet}
\end{subfigure}
\vspace{10pt}\\
\begin{subfigure}{.3\linewidth}
\def \mynet {convnet}
{\includegraphics[trim=0cm 0.75cm 0cm 1.1cm, clip=yes, height=0.6\linewidth]{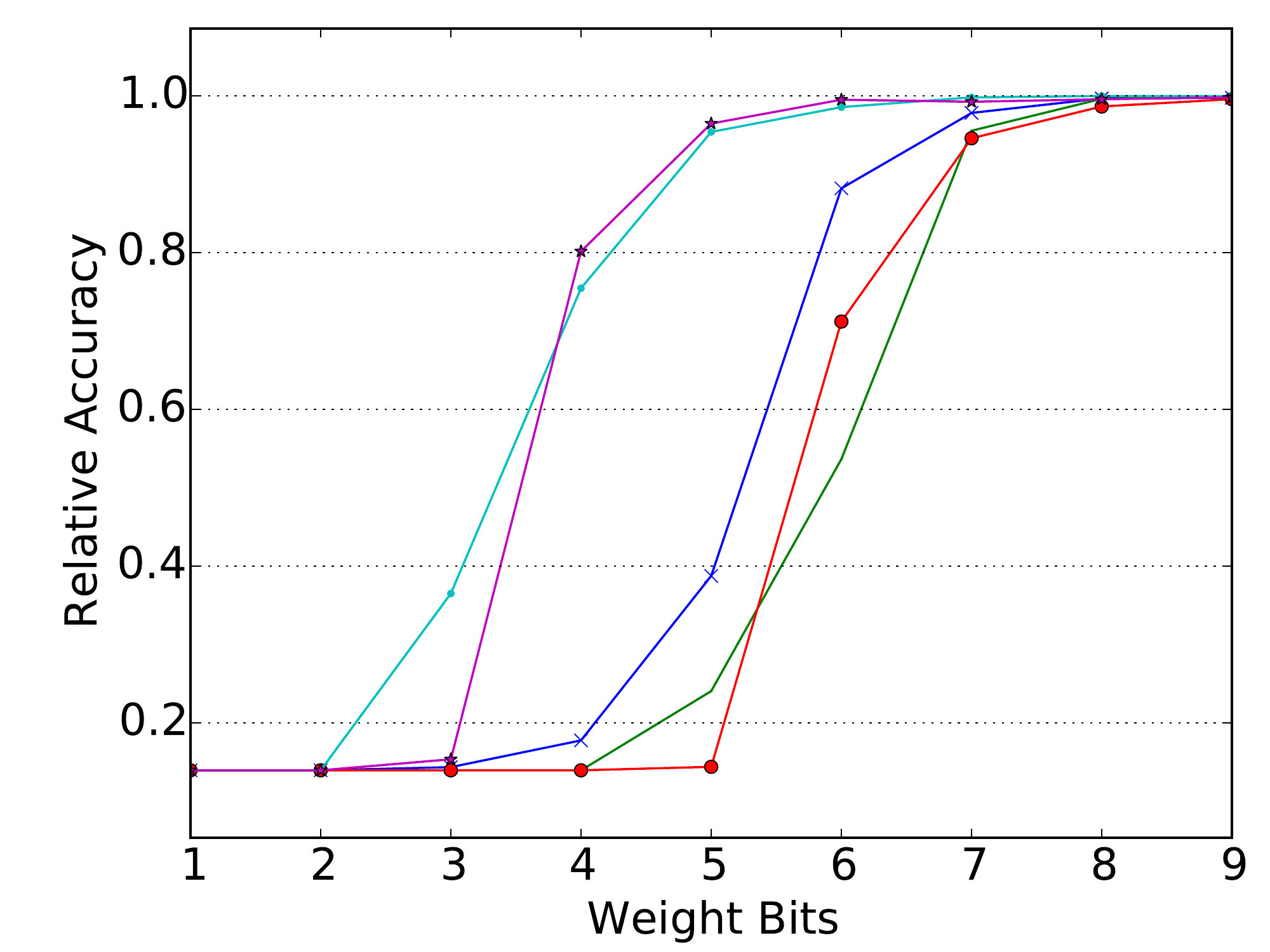}}%
\caption{Convnet: Weights}
\end{subfigure}
\begin{subfigure}{.3\linewidth}
\def \mynet {convnet}
{\includegraphics[trim=0cm 0.75cm 0cm 1.1cm, clip=yes, height=0.6\linewidth]{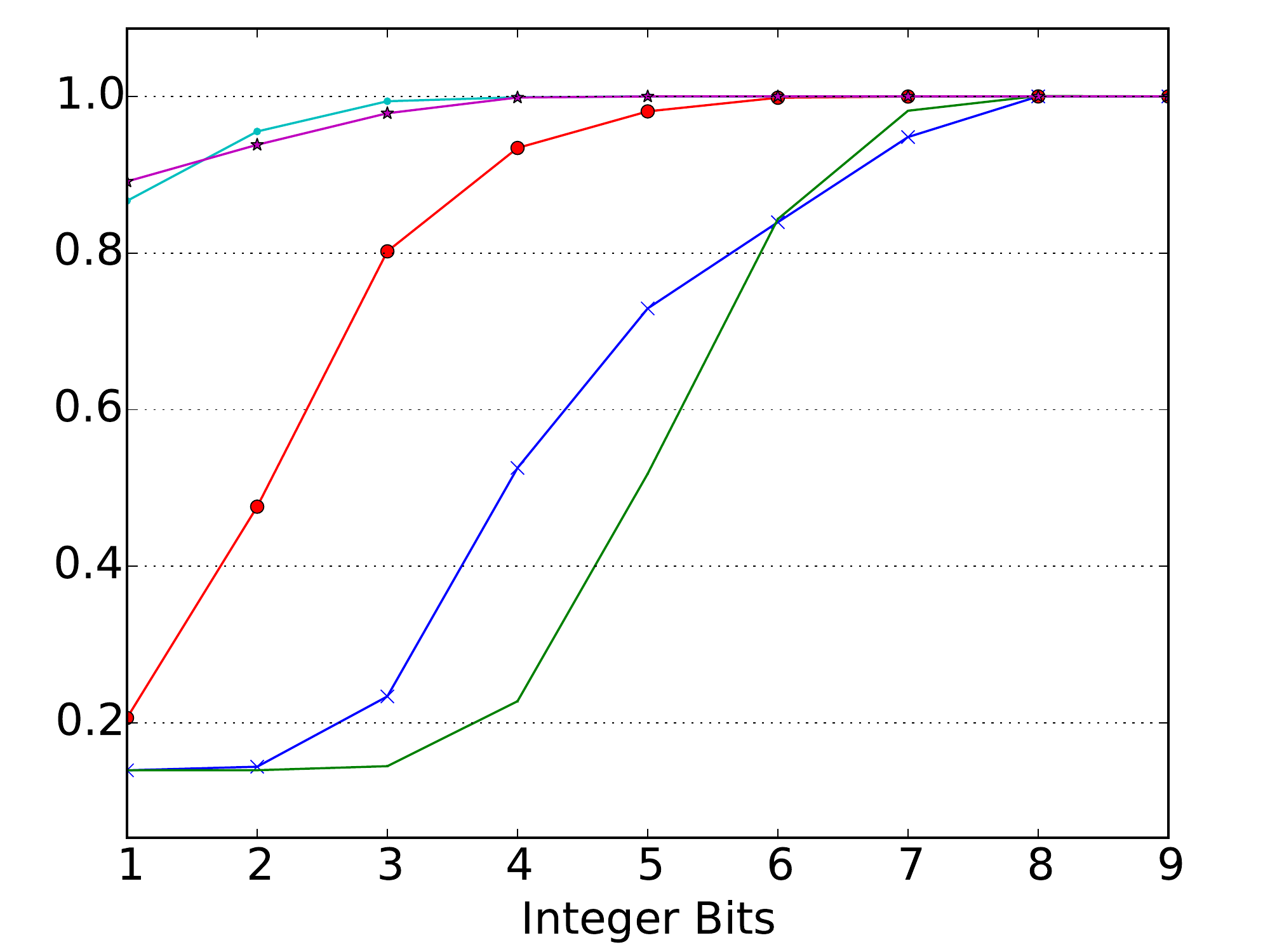}}%
\caption{Convnet: Data: Integer}
\end{subfigure}
\begin{subfigure}{.3\linewidth}
\def \mynet {convnet}
{\includegraphics[trim=0cm 0.75cm 0cm 1.1cm, clip=yes, height=0.6\linewidth]{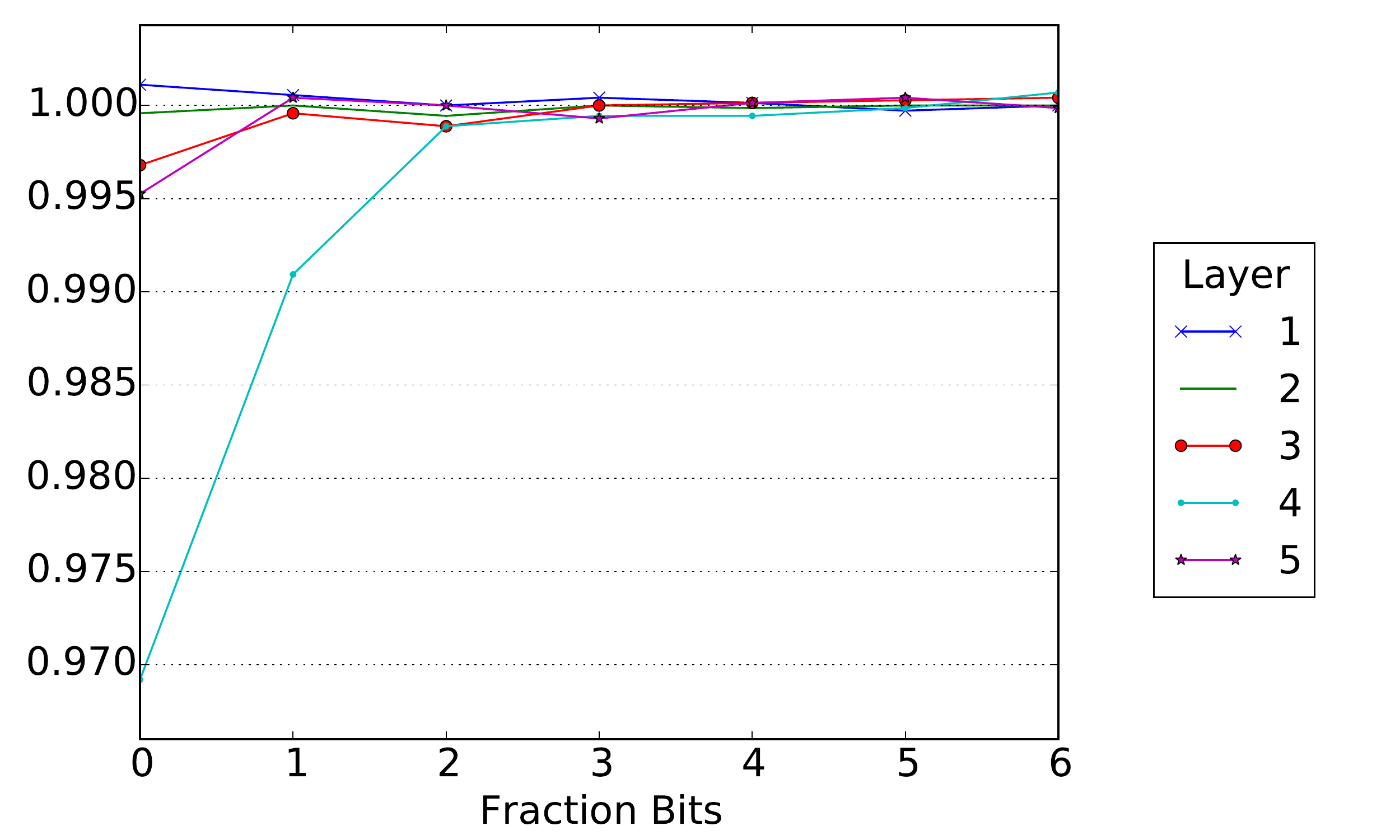}}
\caption{Convnet: Data: Fraction}
\end{subfigure}
\vspace{10pt}\\
\begin{subfigure}{.3\linewidth}
\def \mynet {alexnet}
{\includegraphics[trim=0cm 0.75cm 0cm 1.1cm, clip=yes, height=0.6\linewidth]{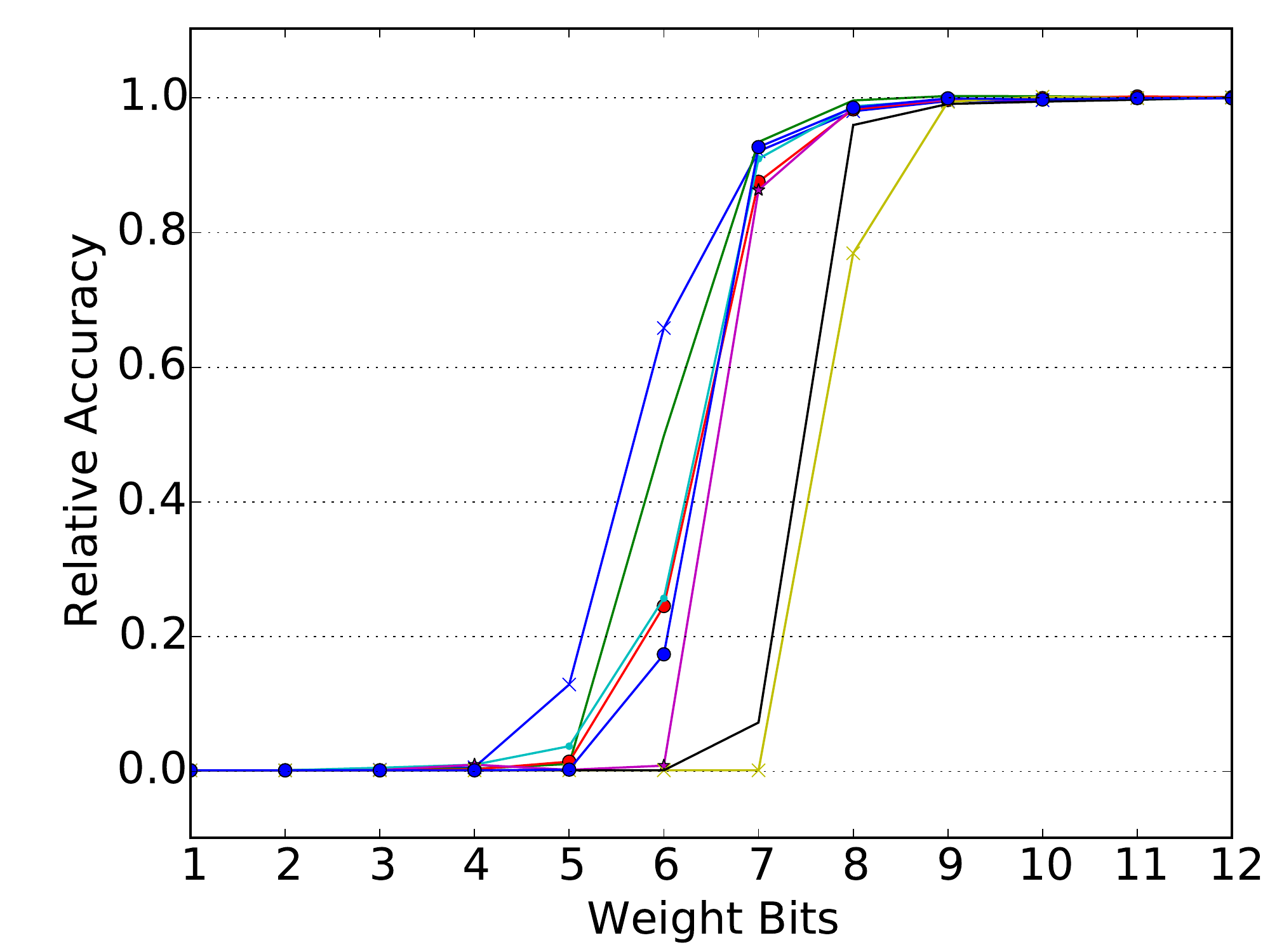}}%
\caption{AlexNet: Weights}
\end{subfigure}
\begin{subfigure}{.3\linewidth}
\def \mynet {alexnet}
{\includegraphics[trim=0cm 0.75cm 0cm 1.1cm, clip=yes, height=0.6\linewidth]{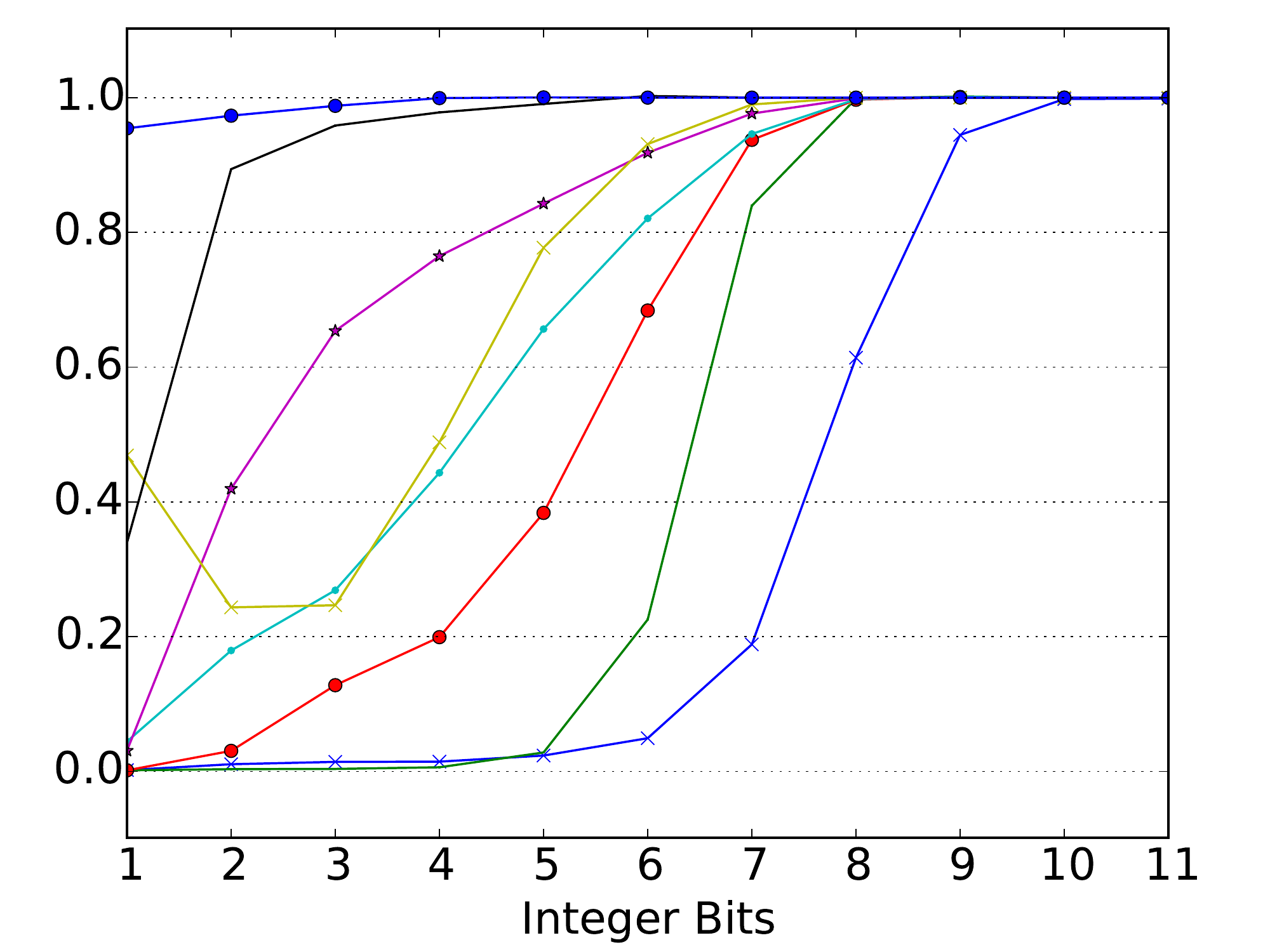}}%
\caption{AlexNet: Data: Integer}
\end{subfigure}
\begin{subfigure}{.3\linewidth}
\def \mynet {alexnet}
{\includegraphics[trim=0cm 0.75cm 0cm 1.1cm, clip=yes, height=0.6\linewidth]{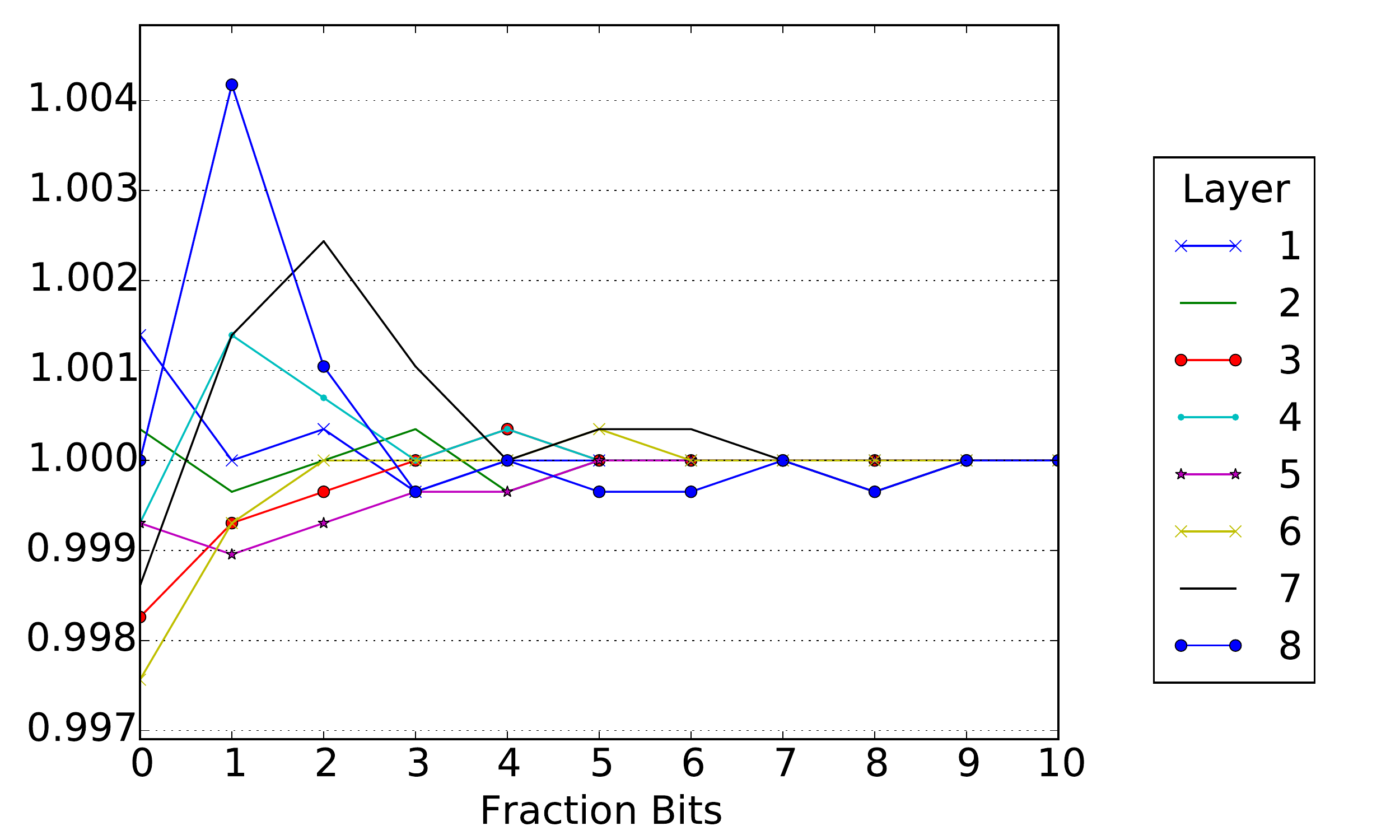}}
\caption{AlexNet: Data: Fraction}
\end{subfigure}
\vspace{10pt}\\
\begin{subfigure}{.3\linewidth}
\def \mynet {nin}
{\includegraphics[trim=0cm 0.75cm 0cm 1.1cm, clip=yes, height=0.6\linewidth]{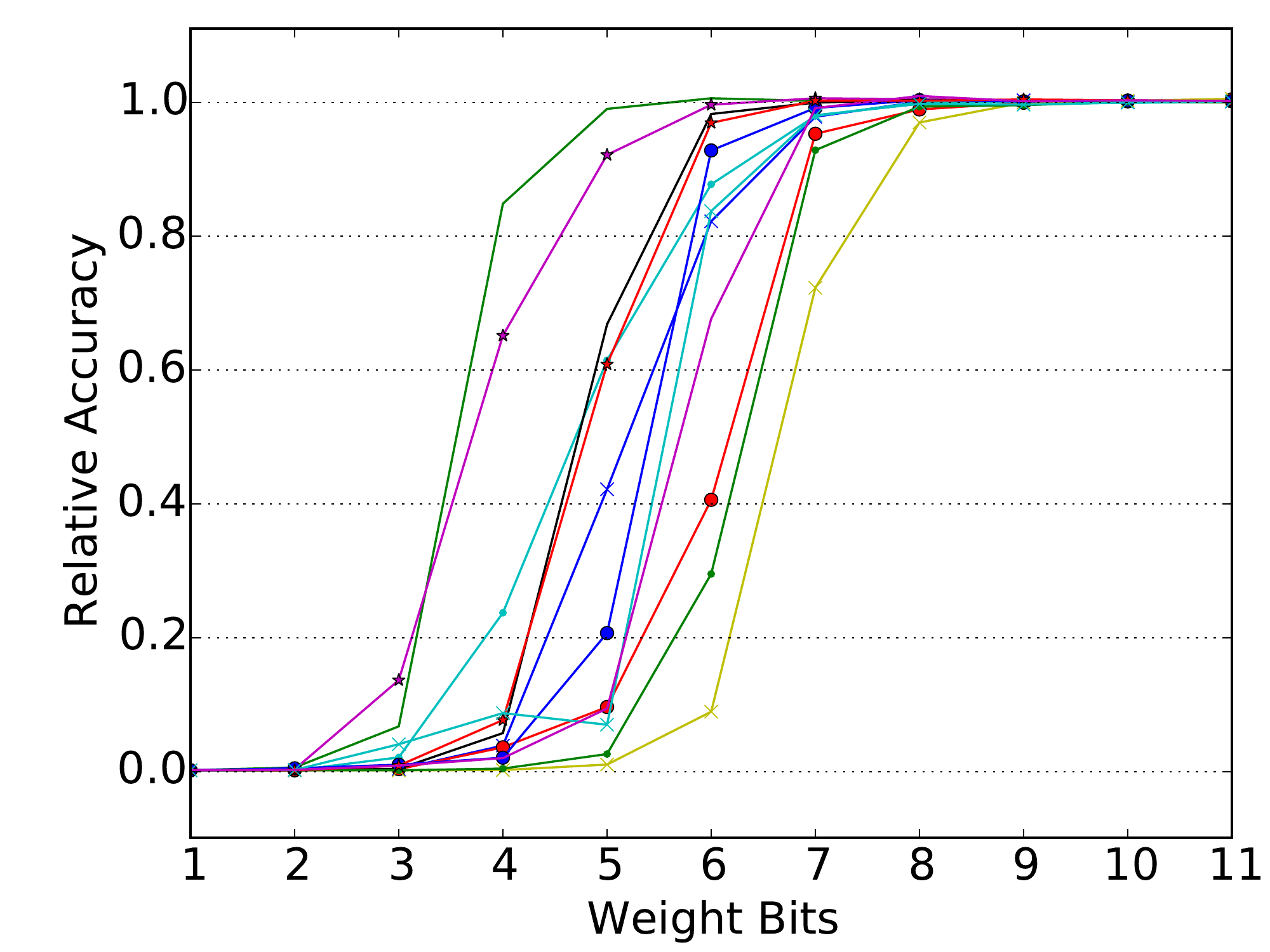}}%
\caption{NiN: Weights}
\end{subfigure}
\begin{subfigure}{.3\linewidth}
\def \mynet {nin}
{\includegraphics[trim=0cm 0.75cm 0cm 1.1cm, clip=yes, height=0.6\linewidth]{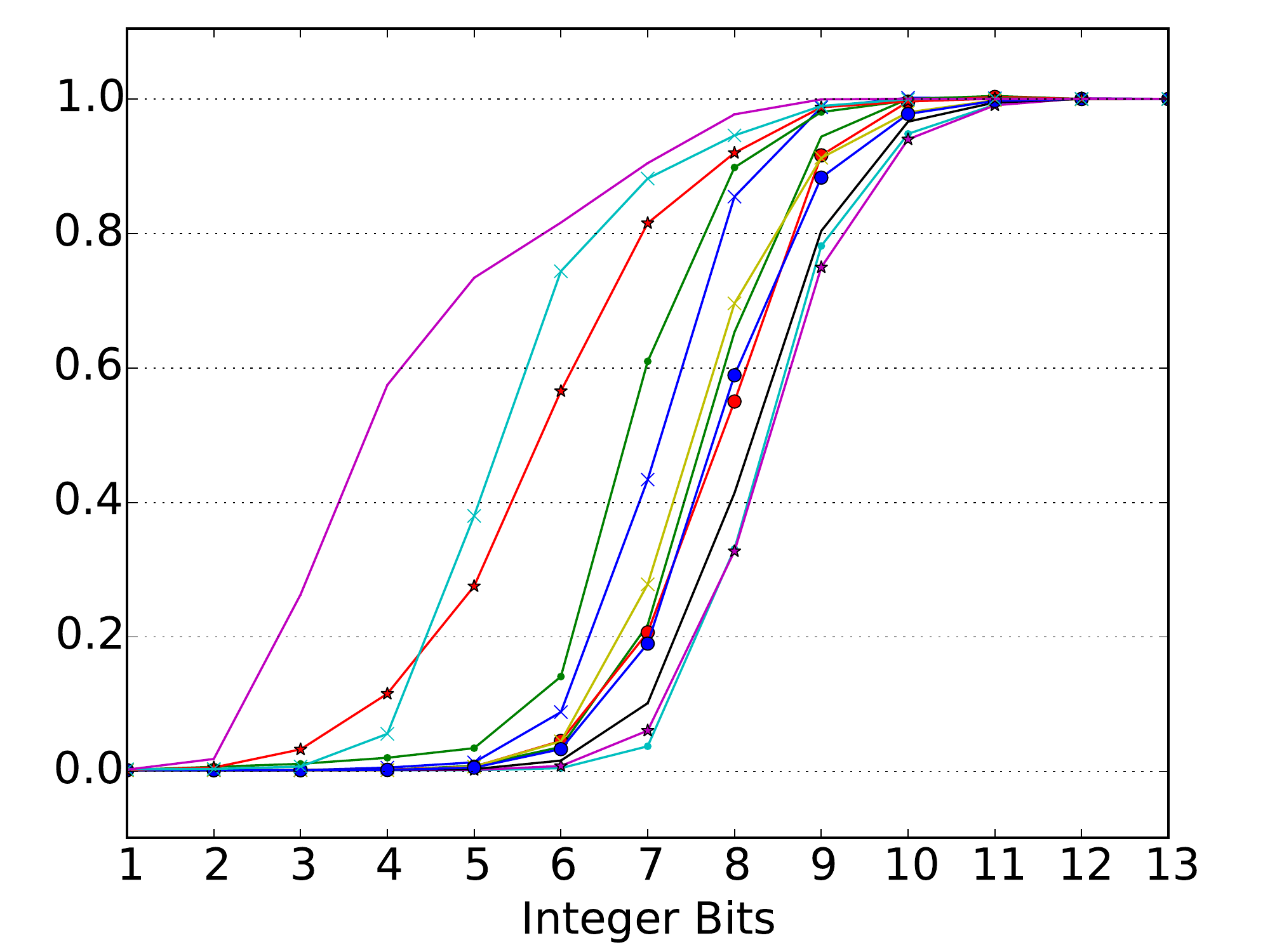}}%
\caption{NiN: Data: Integer}
\end{subfigure}
\begin{subfigure}{.3\linewidth}
\def \mynet {nin}
{\includegraphics[trim=0cm 0.75cm 0cm 1.1cm, clip=yes, height=0.6\linewidth]{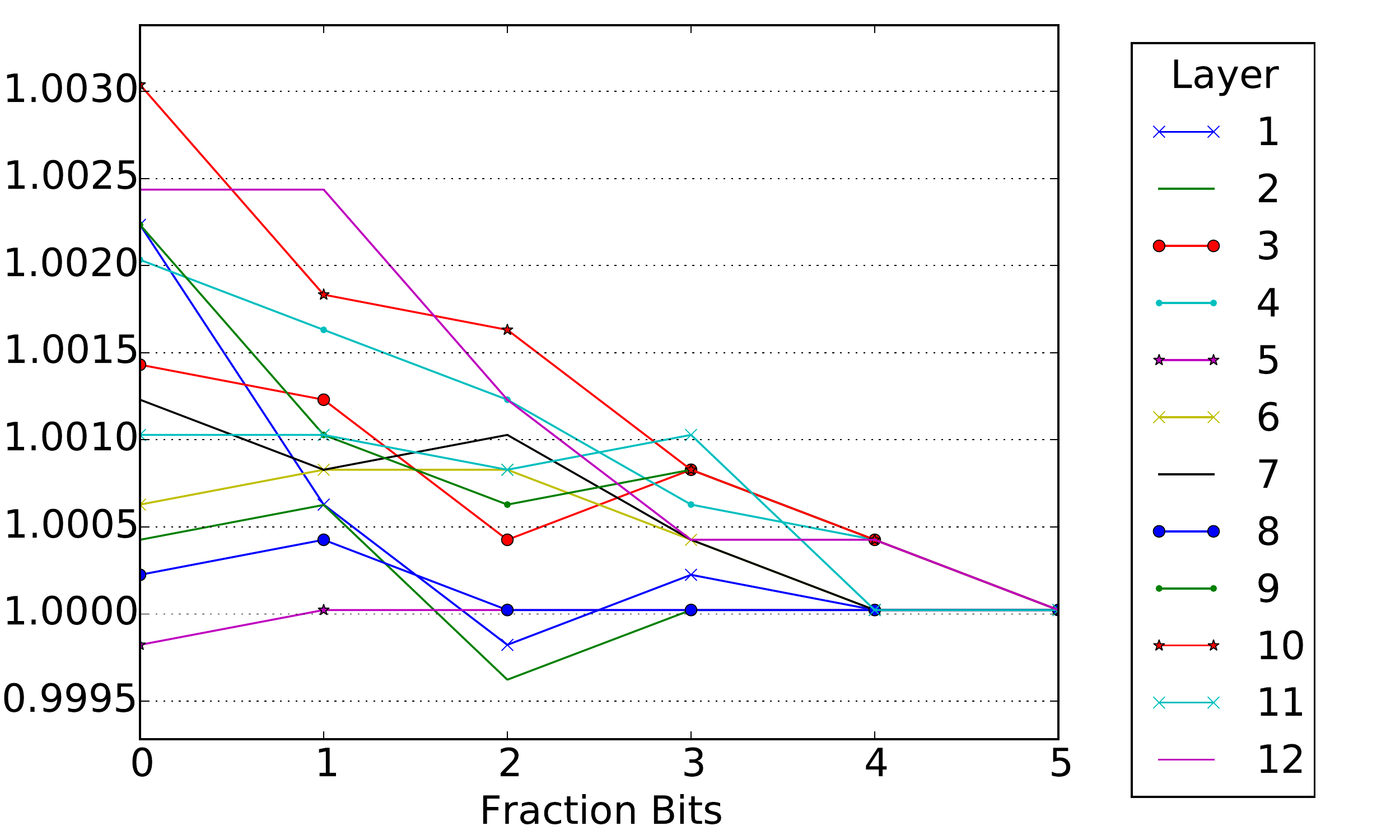}}
\caption{Nin: Data: Fraction}
\end{subfigure}
\vspace{10pt}\\
\begin{subfigure}{.3\linewidth}
\def \mynet {googlenet}
{\includegraphics[trim=0cm 0.75cm 0cm 1.1cm, clip=yes, height=0.6\linewidth]{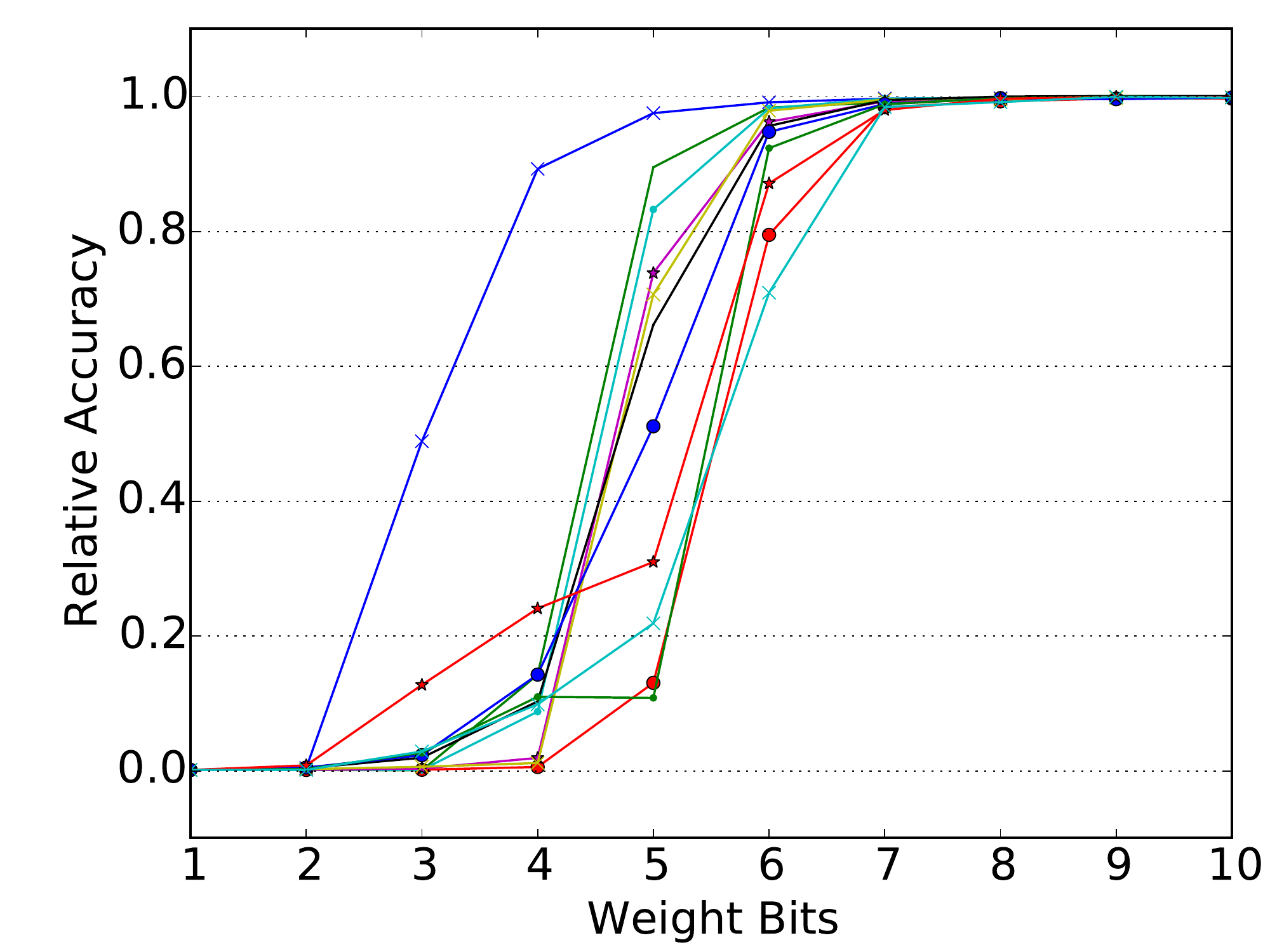}}%
\caption{GoogLeNet: Weights}
\end{subfigure}
\begin{subfigure}{.3\linewidth}
\def \mynet {googlenet}
{\includegraphics[trim=0cm 0.75cm 0cm 1.1cm, clip=yes, height=0.6\linewidth]{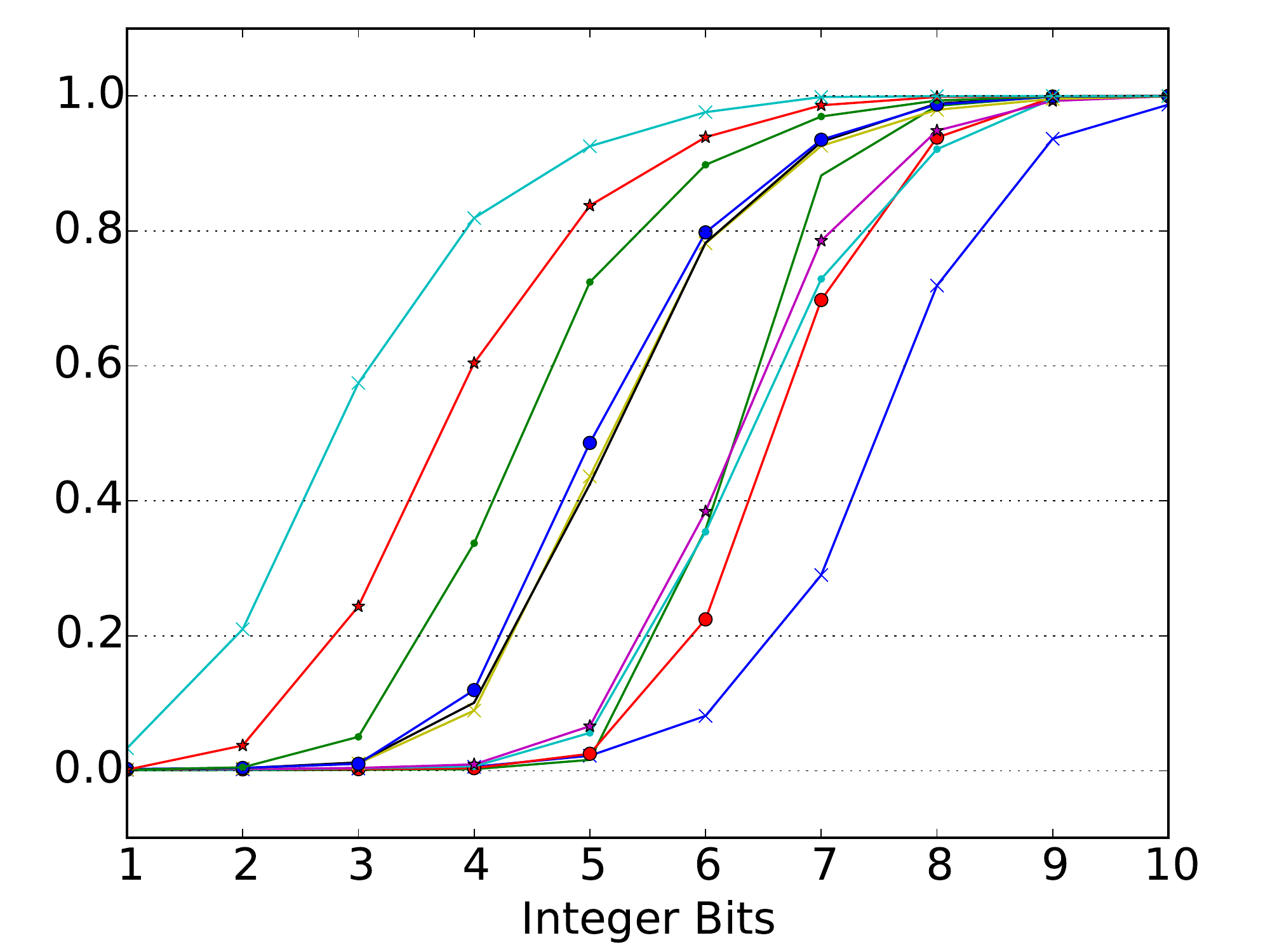}}%
\caption{GoogLeNet: Data: Integer}
\end{subfigure}
\begin{subfigure}{.3\linewidth}
\def \mynet {googlenet}
{\includegraphics[trim=0cm 0.75cm 0cm 1.1cm, clip=yes, height=0.6\linewidth]{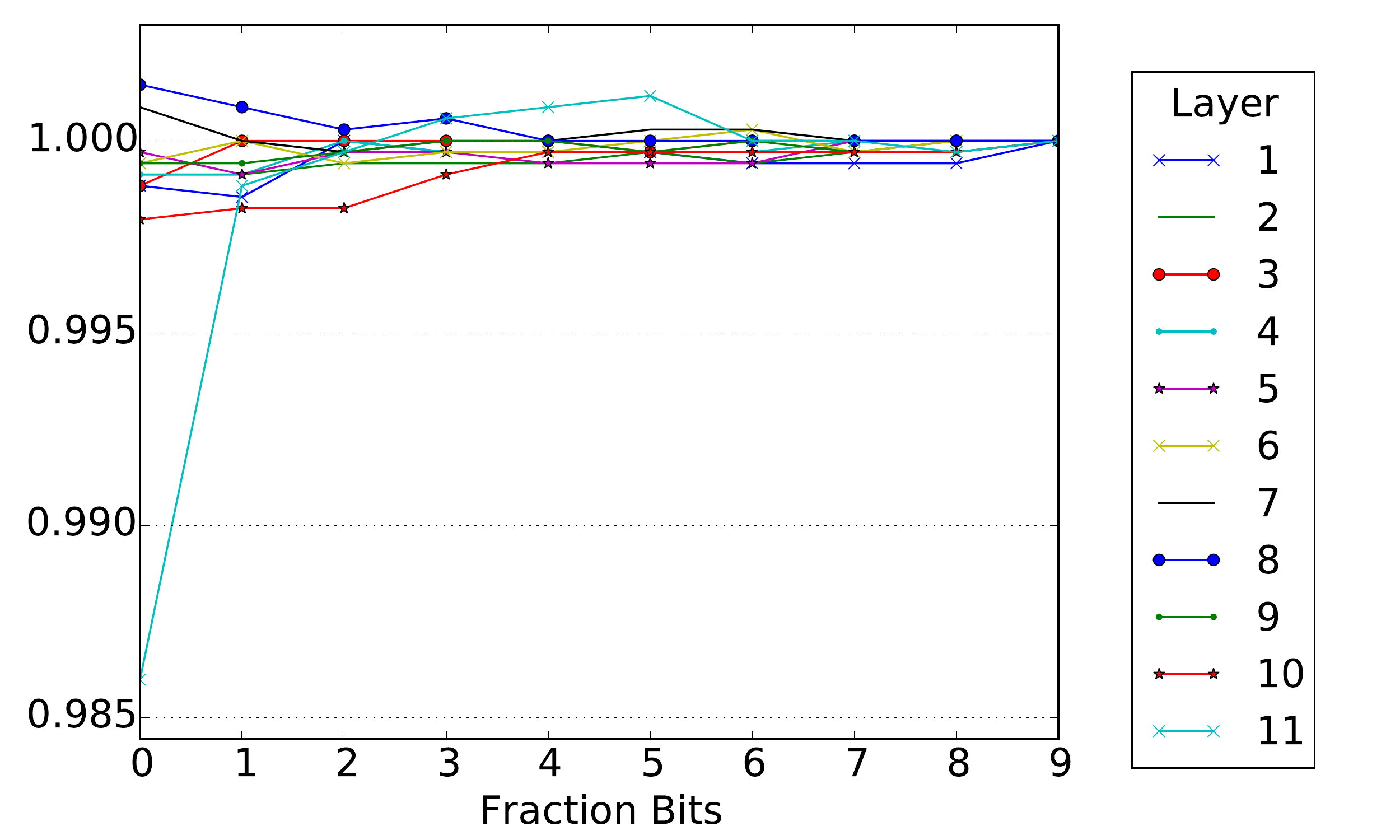}}
\caption{GoogLeNet: Data: Fraction}
\end{subfigure}
\caption{Accuracy vs. Representation Length Per Layer: Weights (left), Intermediate Results: Integer (center), and Fraction (right) portions.
}
\label{fig:perlayer}
\end{figure*}

\subsection{Per Layer Representation Requirements}
\label{sec:perlayerneeds}
\vspace{0.2cm}
This section demonstrates that precision requirements vary even \textit{within} each network. This suggest that we should allow   each layer to use a different representation, reducing memory footprint and communication needs further. For the purpose of these experiments we maintain the baseline numerical representation for all layers, and vary the representation for one layer at a time. Section~\ref{sec:optimal} considers the combined effect of choosing different representations per layer for all layers simultaneously. 

\vspace{0.2cm}
\noindent{\bf Weights:}
The first column of Fig. ~\ref{fig:perlayer} shows how CNN accuracy varies when we change the fixed-point representation used for the weights of one layer at time.
As weights are typically between -1 and 1, we use a single integer bit (sign bit) and vary the number of fractional bits used by the fixed point representation.
The results show that the minimum number of bits needed varies per layer and per network.
For example, in LeNet, three bits are sufficient for layer 2, whereas seven bits are needed for layer 3, and in NiN, five bits are needed for layer 2 while nine are needed for layer 3. 

\vspace{0.2cm}
\noindent{\bf Data:}
The last two columns of Fig. ~\ref{fig:perlayer} show how CNN accuracy varies when we change respectively the integer and the fractional portions of the fixed-point representation used for the intermediate data values one layer at time.
Recall, that the \textit{data} are the inputs and the values produced and communicated between layers.
Focusing on middle column of  Fig. ~\ref{fig:perlayer}, the integer portion requirements vary greatly across layers and across networks.
For example, for Convnet, layer 4 requires just three bits, whereas layer 1 needs eight.
The right column of Fig. ~\ref{fig:perlayer} shows that variation exists in the per layer and per network needs for the fractional portion as well but to a lesser extent for each network.



\subsection{Data Traffic Measurements}
\label{sec:datatraffic}

This section reports the number of data access performed by the networks including the input data, the intermediate data read and written by the layers, and the weights. We will use these measurements in the next section where we present a method for selecting different precisions per layer with the goal of minimizing overall data traffic.

The reported measurements underestimate the amount of traffic generated by the CNNs and thus the benefits that may be possible from reducing off-chip traffic.
Specifically, these experiments assume that once a layer touches a piece of data this data is transferred from or to memory only once for the duration of the layer's execution.
In practice, layers read values multiple times.
These measurements assume that there is enough buffering on chip to capture any data reuse by a layer no matter the reuse distance.
In practice this may not be possible as the amount of buffering needed may be prohibitive.
 For example, a convolution layer will need to buffer multiple full lines or tiles of its input to avoid reading data twice.
 This may be impractical depending on the image size, e.g., for images or video captured by modern mobile devices.

\begin{figure}[ht]
\centering
\captionsetup[subfigure]{oneside,margin={0cm,-2cm}}
\begin{subfigure}{0.48\linewidth}
\includegraphics[trim=-5cm 0.5cm 0cm 0cm, clip=false, height=0.44\textwidth]{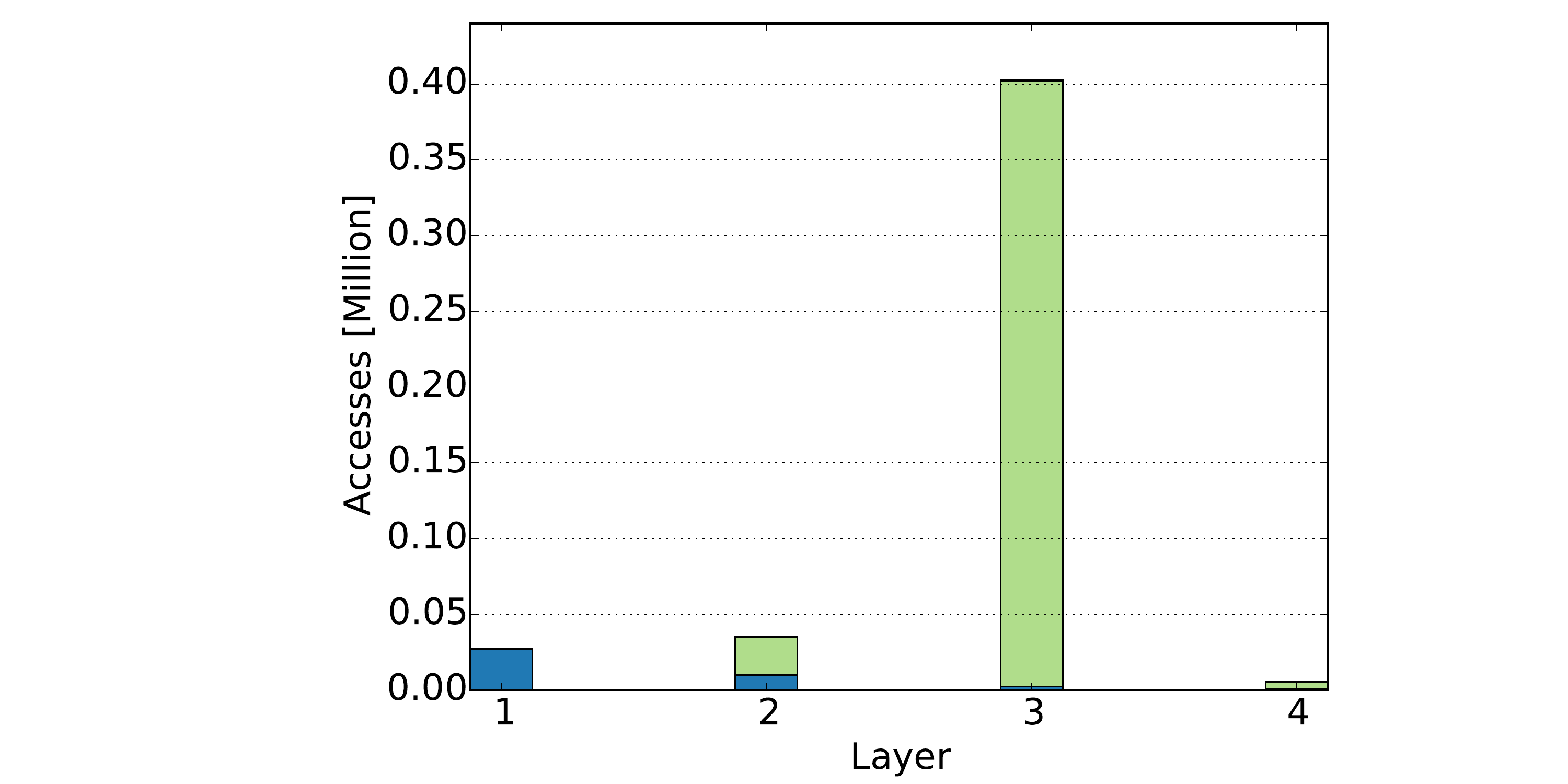} 
\caption{LeNet - Single}
\end{subfigure}
\captionsetup[subfigure]{oneside,margin={0cm,2cm}}
\begin{subfigure}{0.48\linewidth}
\includegraphics[trim=0cm 0.5cm 0cm 0cm, clip=false, height=0.44\textwidth]{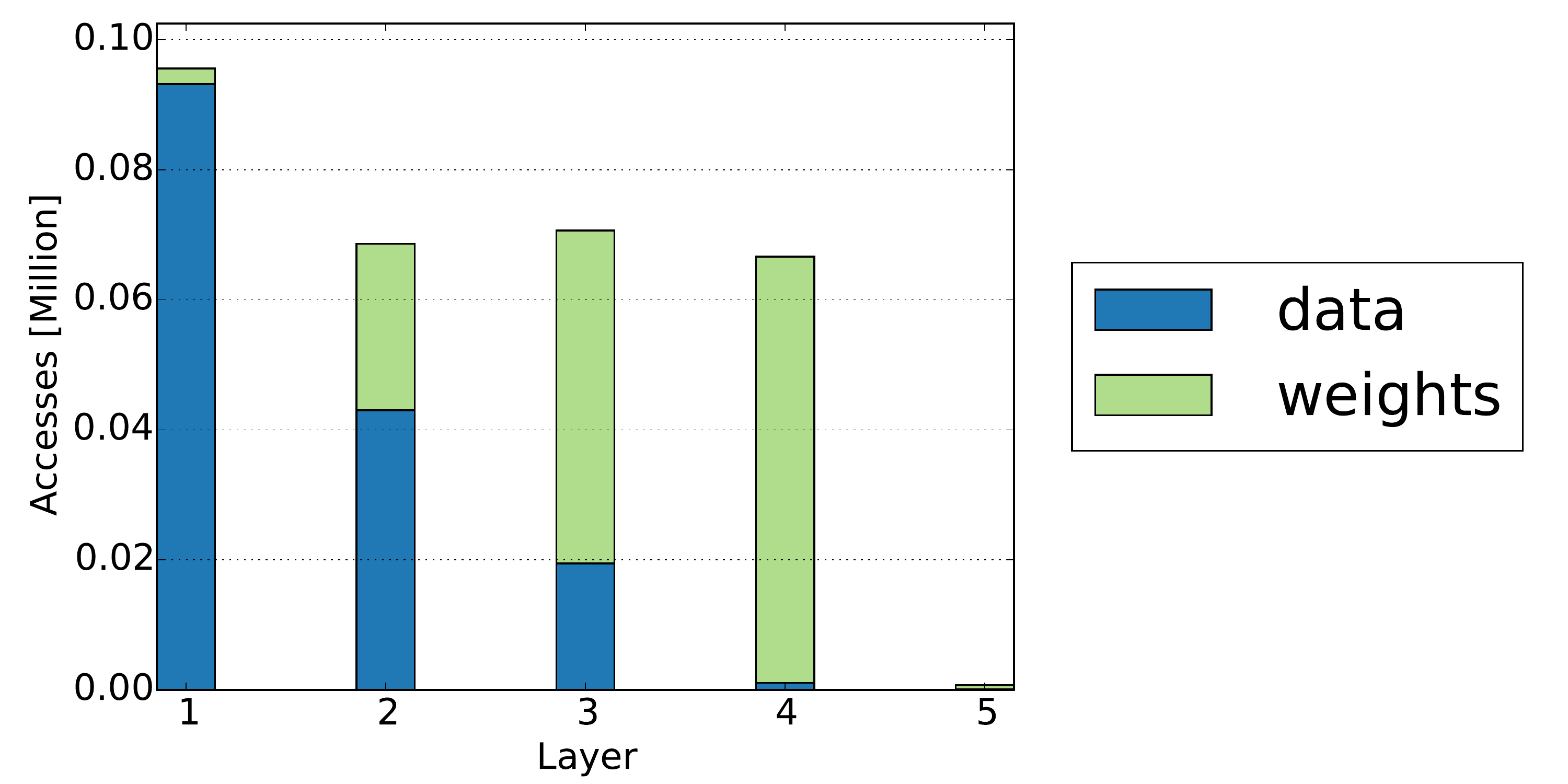}
\caption{Convnet - Single}
\end{subfigure}\\
\captionsetup[subfigure]{oneside,margin={0cm,0cm}}
\vspace{0.2cm}
\begin{subfigure}{0.32\linewidth}
\includegraphics[trim=0cm 0.5cm 2cm 0cm, clip=false, height=0.66\textwidth]{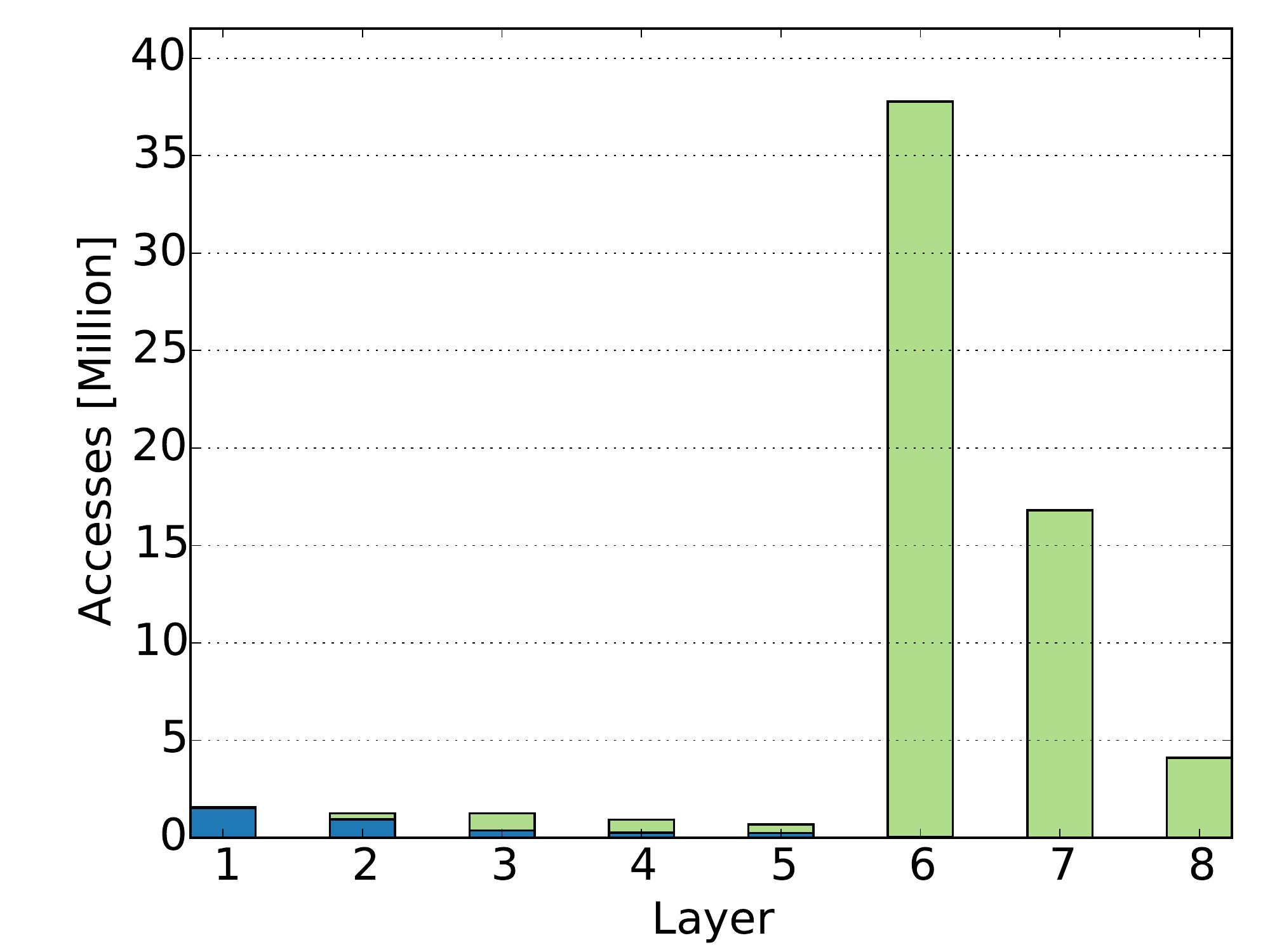}
\caption{Alexnet - Single}
\end{subfigure}
\begin{subfigure}{0.32\linewidth}
\includegraphics[trim=0cm 0.5cm 0cm 0cm, clip=false, height=0.66\textwidth]{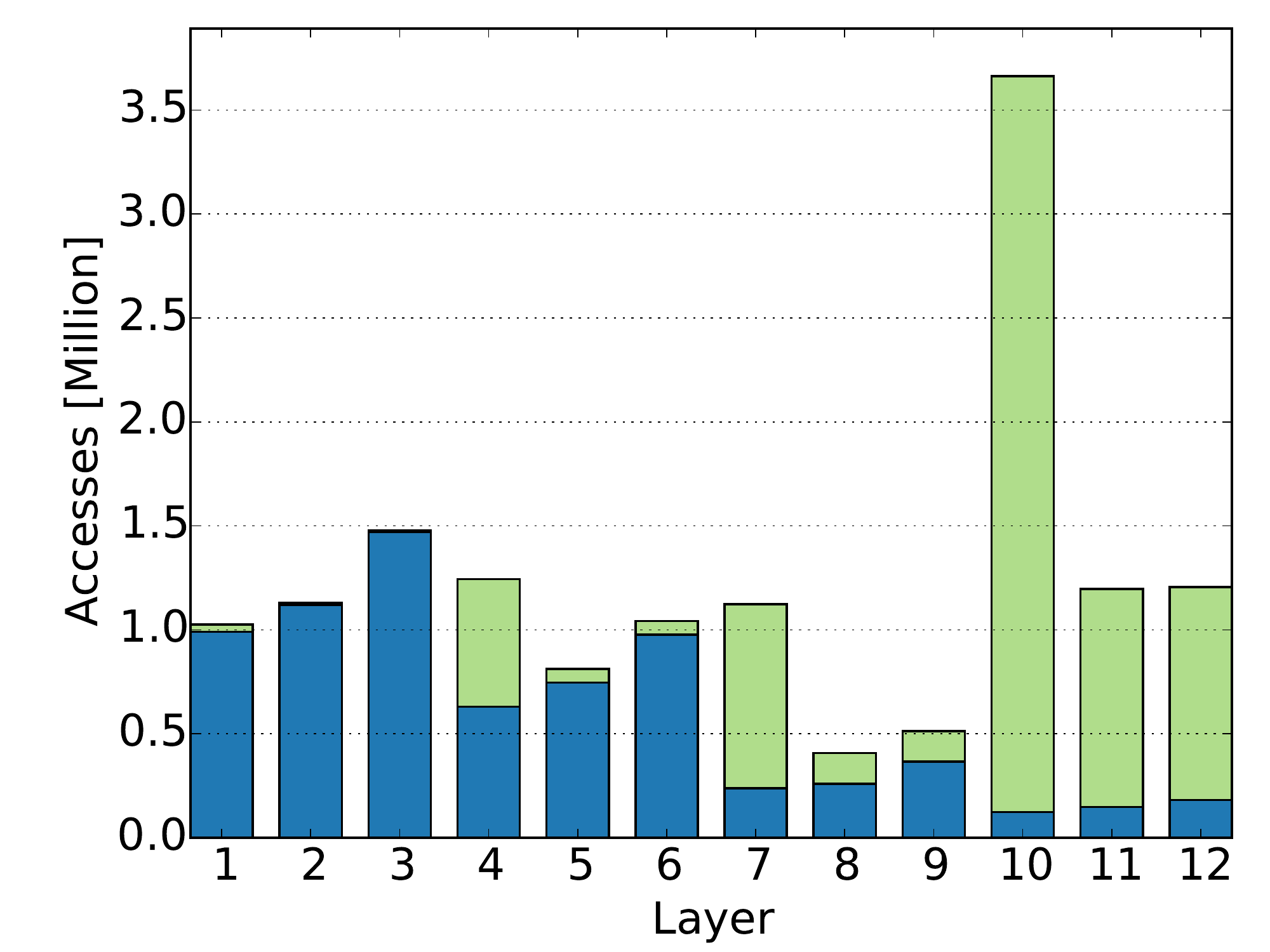}
\caption{NiN - Single}
\end{subfigure}
\begin{subfigure}{0.32\linewidth}
\includegraphics[trim=0cm 0.5cm 0cm 0cm, clip=false, height=0.66\textwidth]{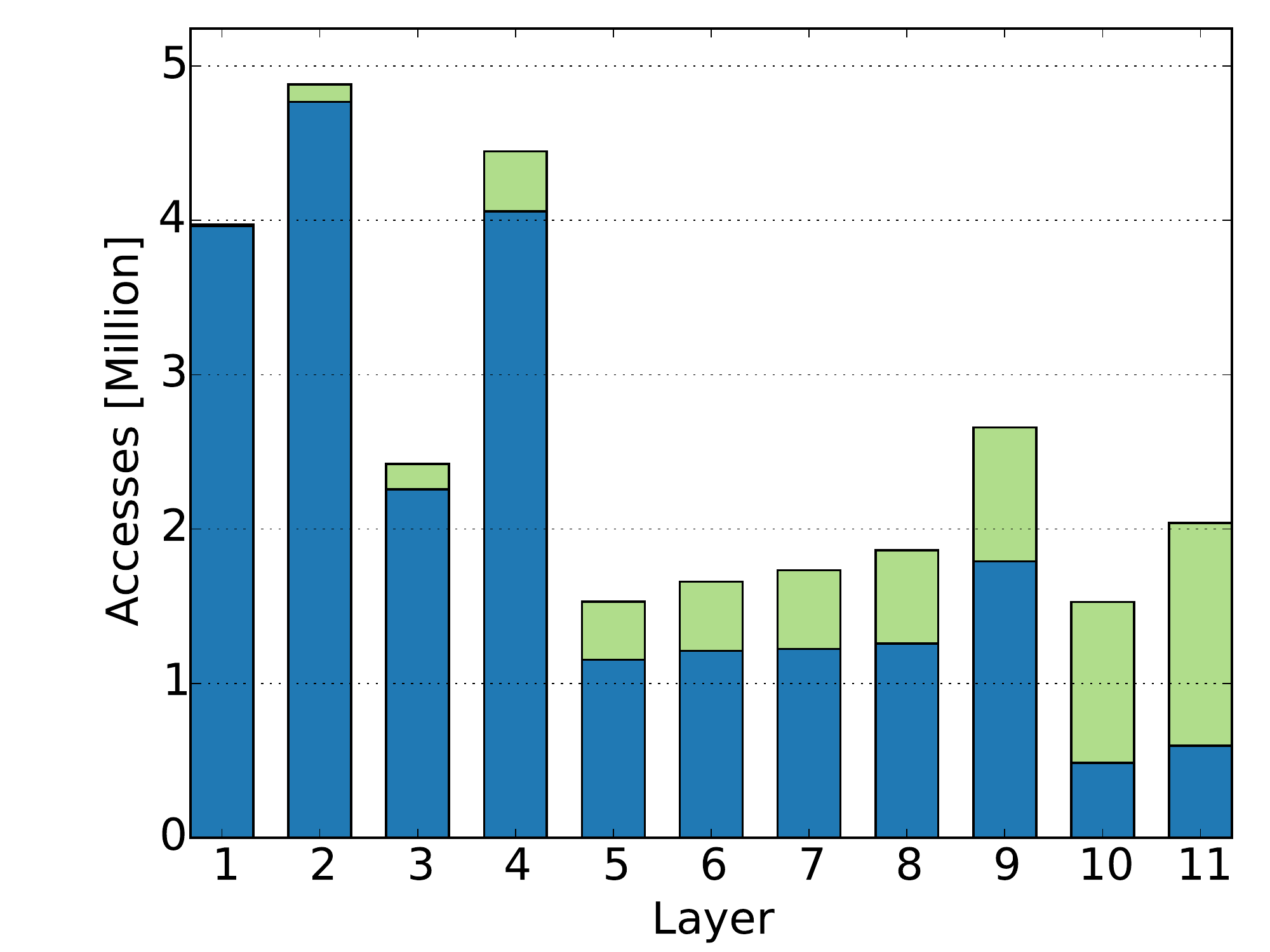}
\caption{GoogLeNet - Single}
\end{subfigure} \\
\vspace{0.2cm}
\captionsetup[subfigure]{oneside,margin={0cm,-2cm}}
\begin{subfigure}{0.48\linewidth}
\includegraphics[trim=-5cm 0.5cm 0cm 0cm, clip=false, height=0.44\textwidth]{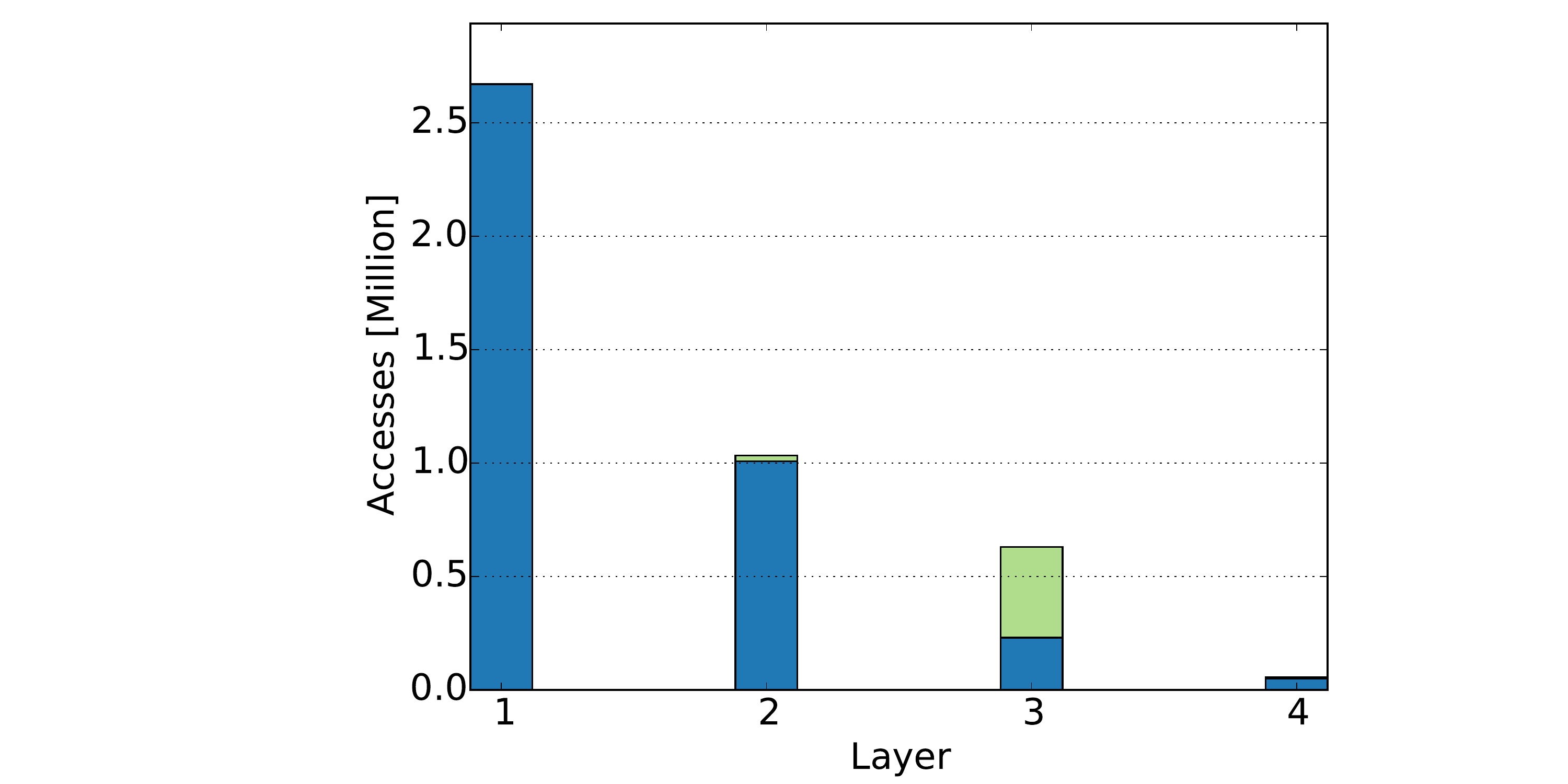}
\caption{LeNet - Batch}
\end{subfigure}
\captionsetup[subfigure]{oneside,margin={0cm,2cm}}
\begin{subfigure}{0.48\linewidth}
\includegraphics[trim=0cm 0.5cm 0cm 0cm, clip=false, height=0.44\textwidth]{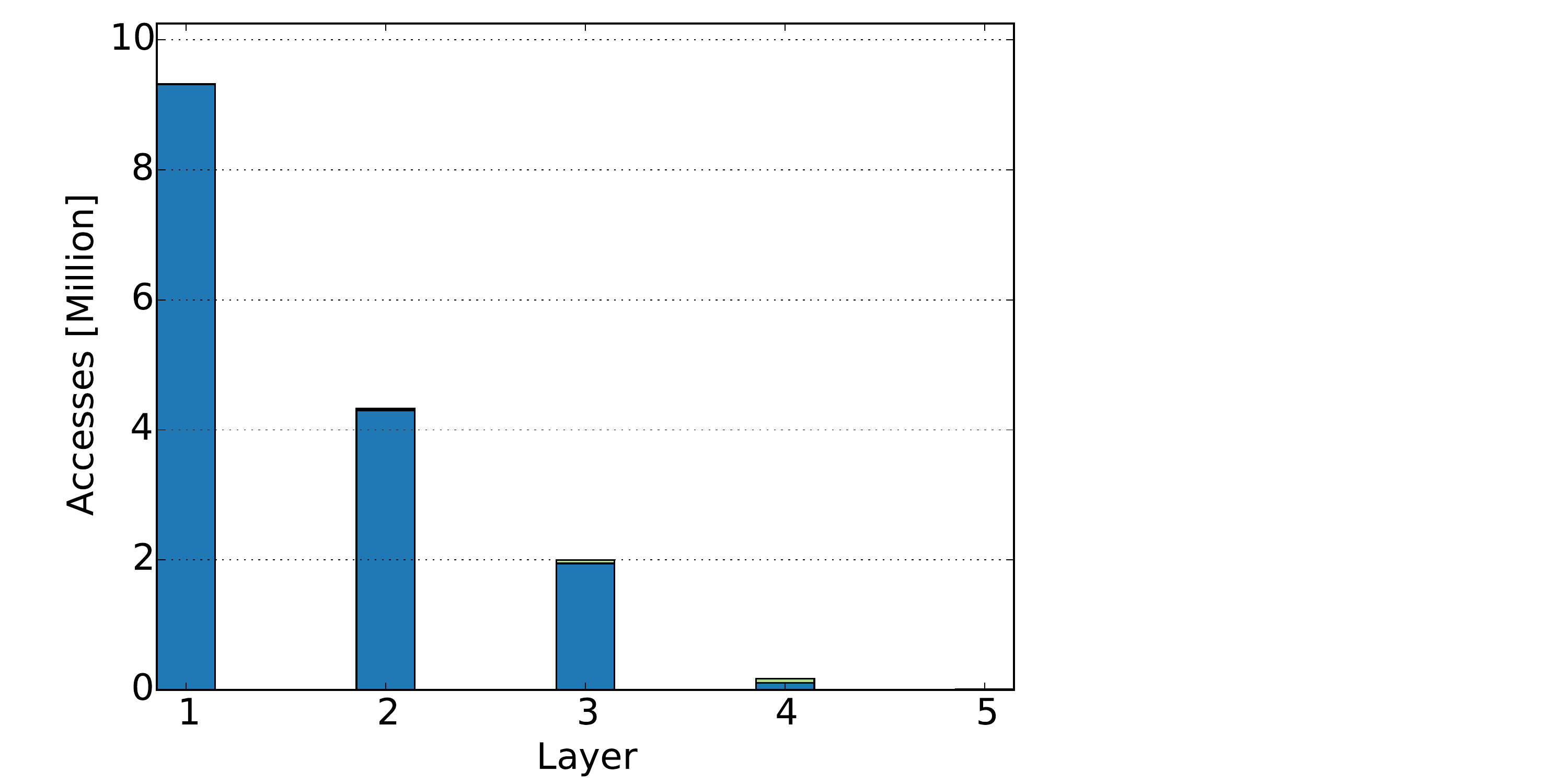}
\caption{Convnet - Batch}
\end{subfigure}\\
\vspace{0.2cm}
\captionsetup[subfigure]{oneside,margin={0cm,0cm}}
\begin{subfigure}{0.32\linewidth}
\includegraphics[trim=0cm 0.5cm 0cm 0cm, clip=false, height=0.66\textwidth]{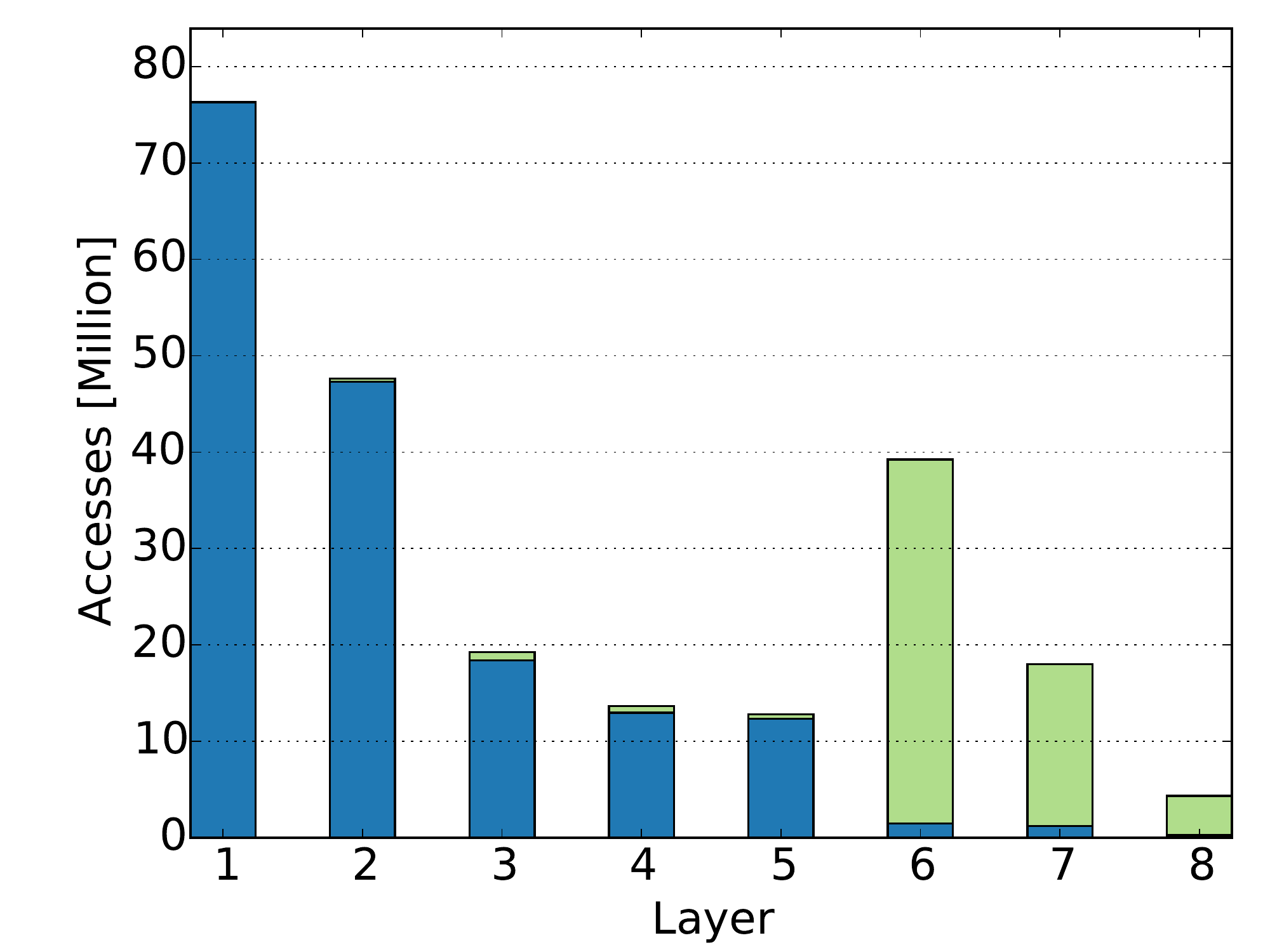}
\caption{Alexnet - Batch}
\end{subfigure}
\begin{subfigure}{0.32\linewidth}
\includegraphics[trim=0cm 0.5cm 0cm 0cm, clip=false, height=0.66\textwidth]{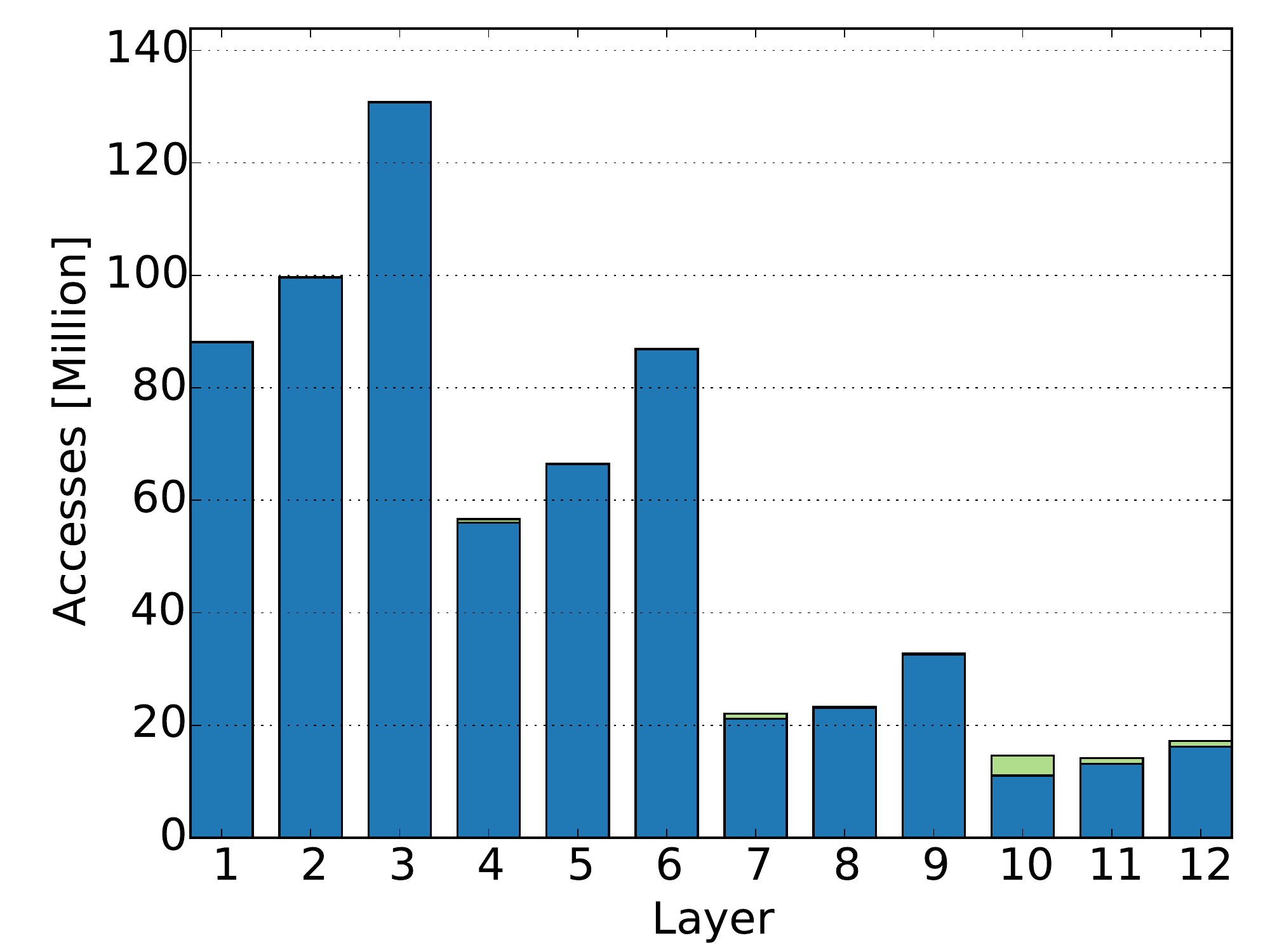}
\caption{NiN - Batch}
\end{subfigure}
\begin{subfigure}{0.32\linewidth}
\includegraphics[trim=0cm 0.5cm 0cm 0cm, clip=false, height=0.66\textwidth]{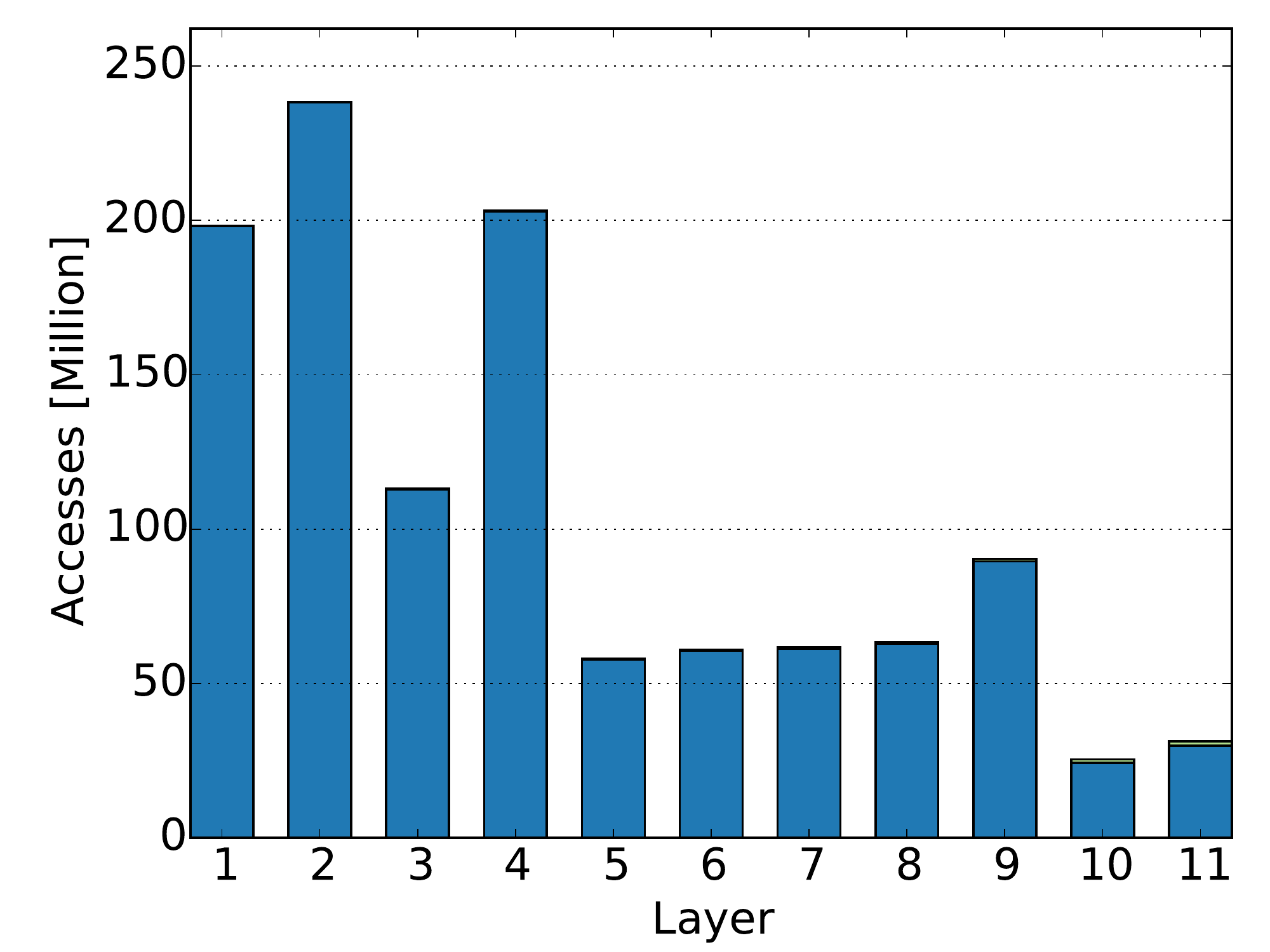}
\caption{GoogLeNet - Batch}
\end{subfigure}
\caption{Data traffic}
\label{fig:data-accesses}
\end{figure}

Fig. \ref{fig:data-accesses} shows the traffic in millions of accesses, where each access is a single data element.
The figure shows two use cases for each network: performing classification on a \textit{single} image or in \textit{batches} of multiple images.
When processing a single image the weights make up a significant portion of the traffic, and dominate traffic in three of the networks.
GoogLeNet is the exception where data dominate traffic even in the single image use case.
In batch processing, the intermediate data dominate traffic.
Batch processing feeds multiple images through the same layer before proceeding with the next layer.
In turn this requires reading the weights from memory only once per layer and not once per image. 
In the single image use case, data dominate at the beginning of the network whereas weights dominate towards the end.
This is expected as  networks tend to use an initial series of convolution layers followed by multiple fully connected layers. 

In the interest of space, and given that when possible batching is used in practice since it reduces how often weights ought to be read, the rest of this work focuses on the batch use case. 

The amount of data that is needed when processing CNNs in practice may potentially be higher depending on how the intermediate computations are performed. 
Also, as image resolution as well as  network fidelity and complexity increases, memory traffic is bound to increase. 

\subsection{Choosing the Per Layer Data Representation} 
\label{sec:optimal}

So far we studied the effect of changing the representation for one layer at a time.
While some loss in fidelity in one layer can be acceptable, this does not suggest that we can simultaneously change the representation for multiple layers and still maintain overall network accuracy.
This section studies how accuracy varies as we adjust the representation used by all layers at the same time.
The goal is to find the \textit{minimum} length representation possible per layer  while maintaining overall network accuracy within acceptable limits.
Such a configuration minimizes data traffic while maintaining accuracy.

To explore the design space in a reasonable amount of time we use gradient descent targeting the output accuracy of the network while adjust the representation length used at each  layer.
Since the reduced precision error tolerance does not strictly decrease as data propagates through the network we cannot easily prune the search space.
For example in Figure \ref{fig:perlayer}(h) AlexNet layer 5 is more error tolerant than layer 6.
As such it is necessary to explore an exponential space of configurations to consider.
To make the search tractable we use slowest gradient descent to approximate the Pareto frontier in the accuracy vs. data traffic space.
The algorithm is as follows:

\begin{enumerate}
\item Initialize all layers to a uniform precision with less that 0.1\% error, found in Figure \ref{fig:4nets-uniform}(a)
\item Create a set of delta configurations by reducing each parameter, integer bits and fractional bits, in each layer by one.
\item Use the delta configuration with the best accuracy to initialize the next iteration.
\end{enumerate}

For the more complex networks, AlexNet, NiN and GoogLeNet, running this iterative algorithm is time consuming.
To reduce the parameter space we fix the fractional bits to 0, 0 and 2 respectively.
These bit values achieve less than 0.1\% error in Figure~\ref{fig:perlayer} (right column).
LeNet and Convnet are simpler networks and are less tolerant to fractional error so we also vary the fractional bits in our exploration for these networks.

\begin{figure}[t]
\centering
\begin{subfigure}{0.32\linewidth}
    {\includegraphics[trim=0cm 1cm 0cm 0.0cm, clip=yes, width=0.99\textwidth]{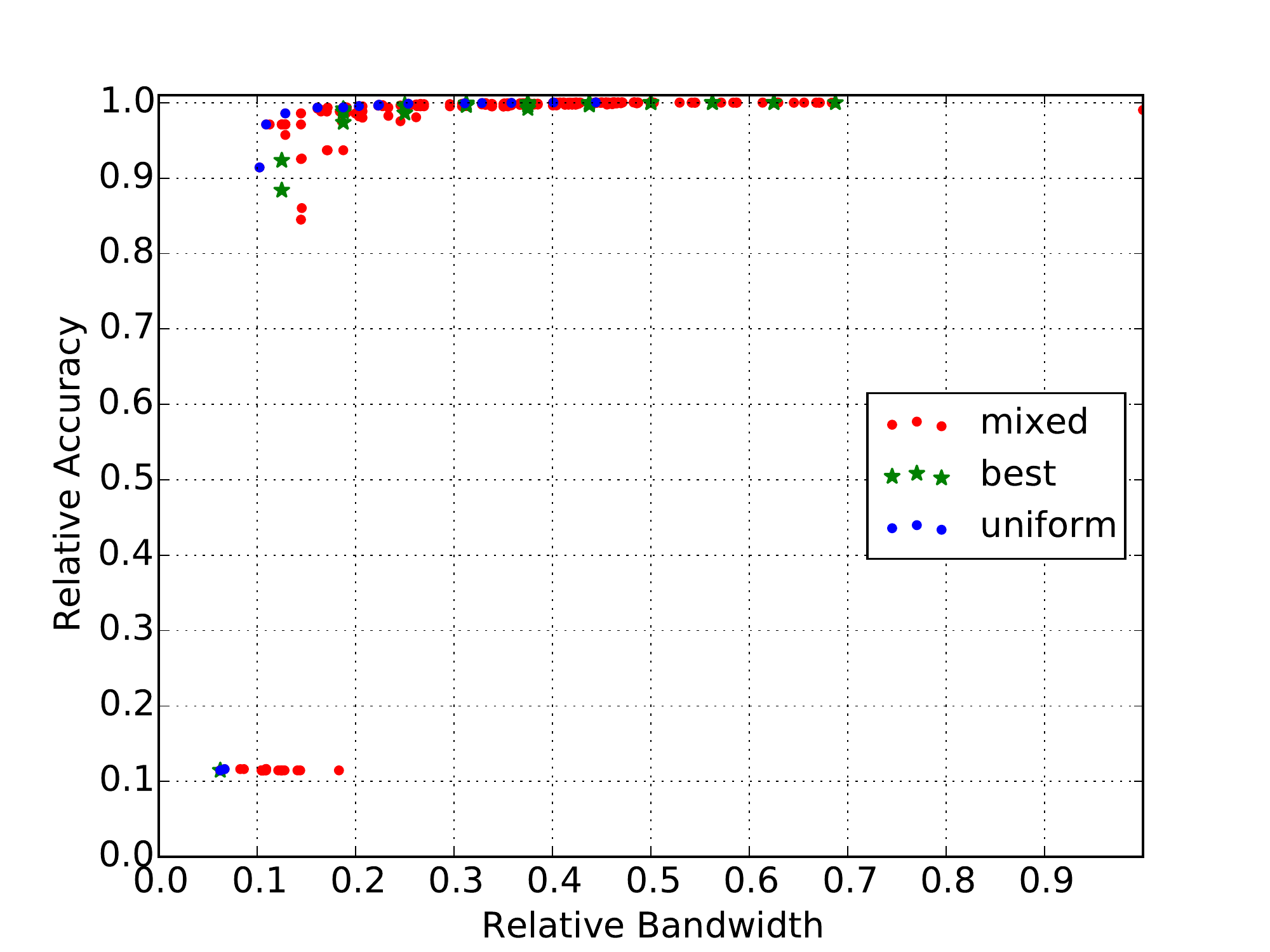}}\\
    \vspace{-0.3cm}
    \caption{LeNet}
\end{subfigure}
\begin{subfigure}{0.32\linewidth}
 {\includegraphics[trim=0cm 1cm 0cm 0.0cm, clip=yes, width=0.99\textwidth]{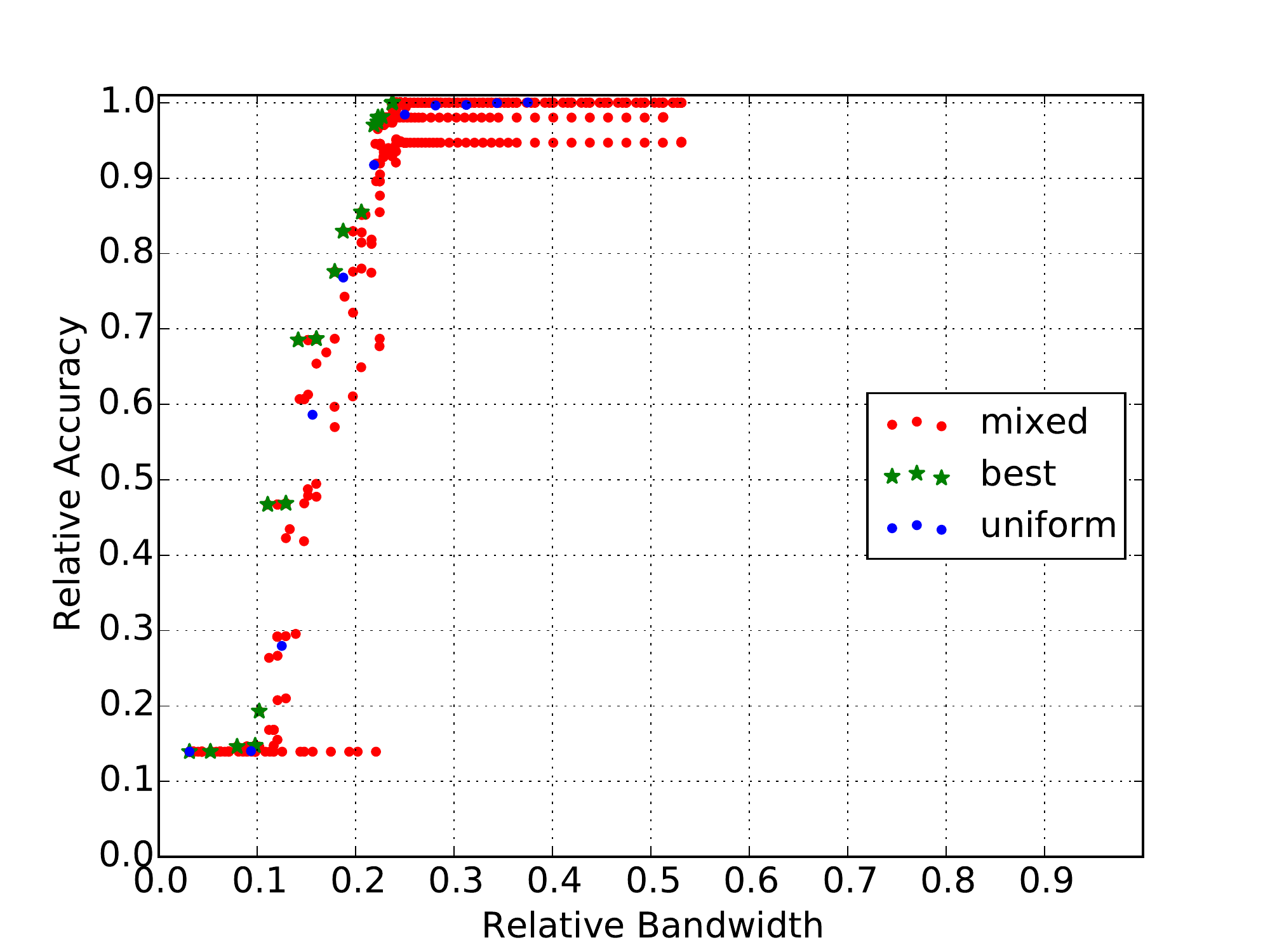}}\\
 \vspace{-0.3cm}\caption{Convnet} 
\end{subfigure}
\begin{subfigure}{0.32\linewidth}
 {\includegraphics[trim=0cm 1cm 0cm 0.0cm, clip=yes, width=0.99\textwidth]{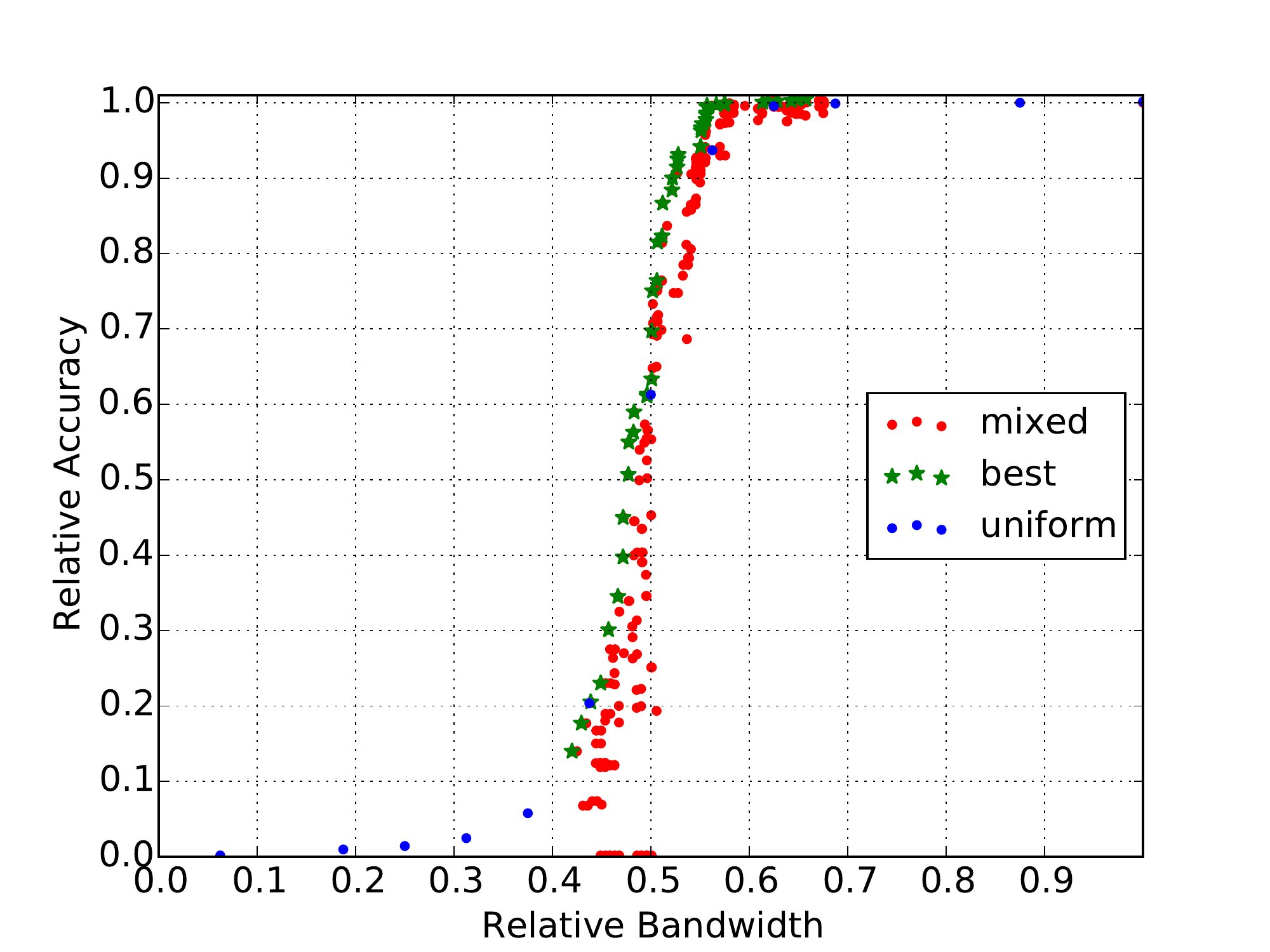}}\\
\vspace{-0.3cm} \caption{AlexNet} 
\end{subfigure}
\begin{subfigure}{0.32\linewidth}
      {\includegraphics[trim=0cm 1cm 0cm 0.0cm, clip=yes, width=0.99\textwidth]{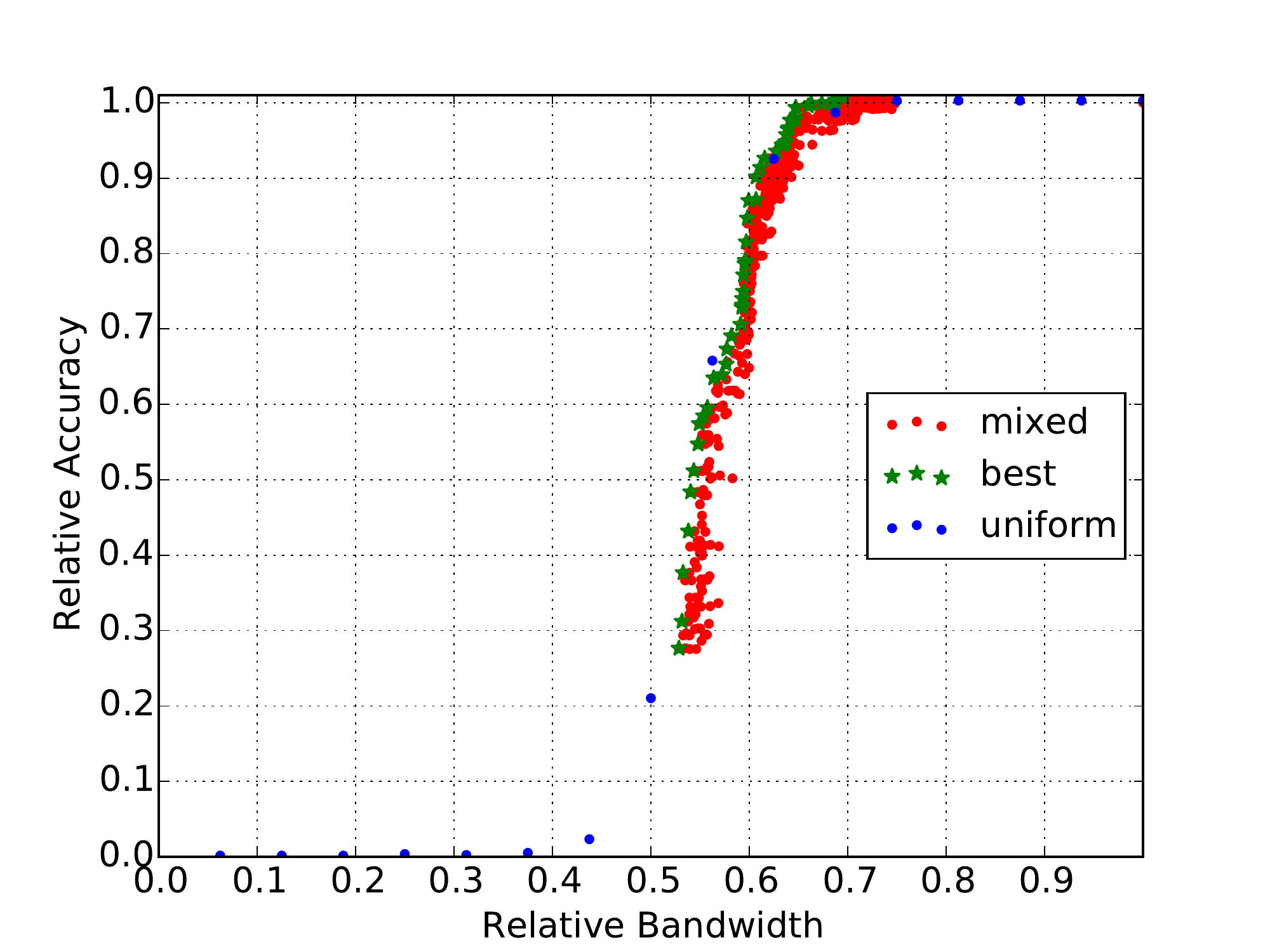}}\\ 
\vspace{-0.3cm}      \caption{NiN}
\end{subfigure}
\begin{subfigure}{0.32\linewidth}
{\includegraphics[trim=0cm 1cm 0cm 0.0cm, clip=yes, width=0.99\textwidth]{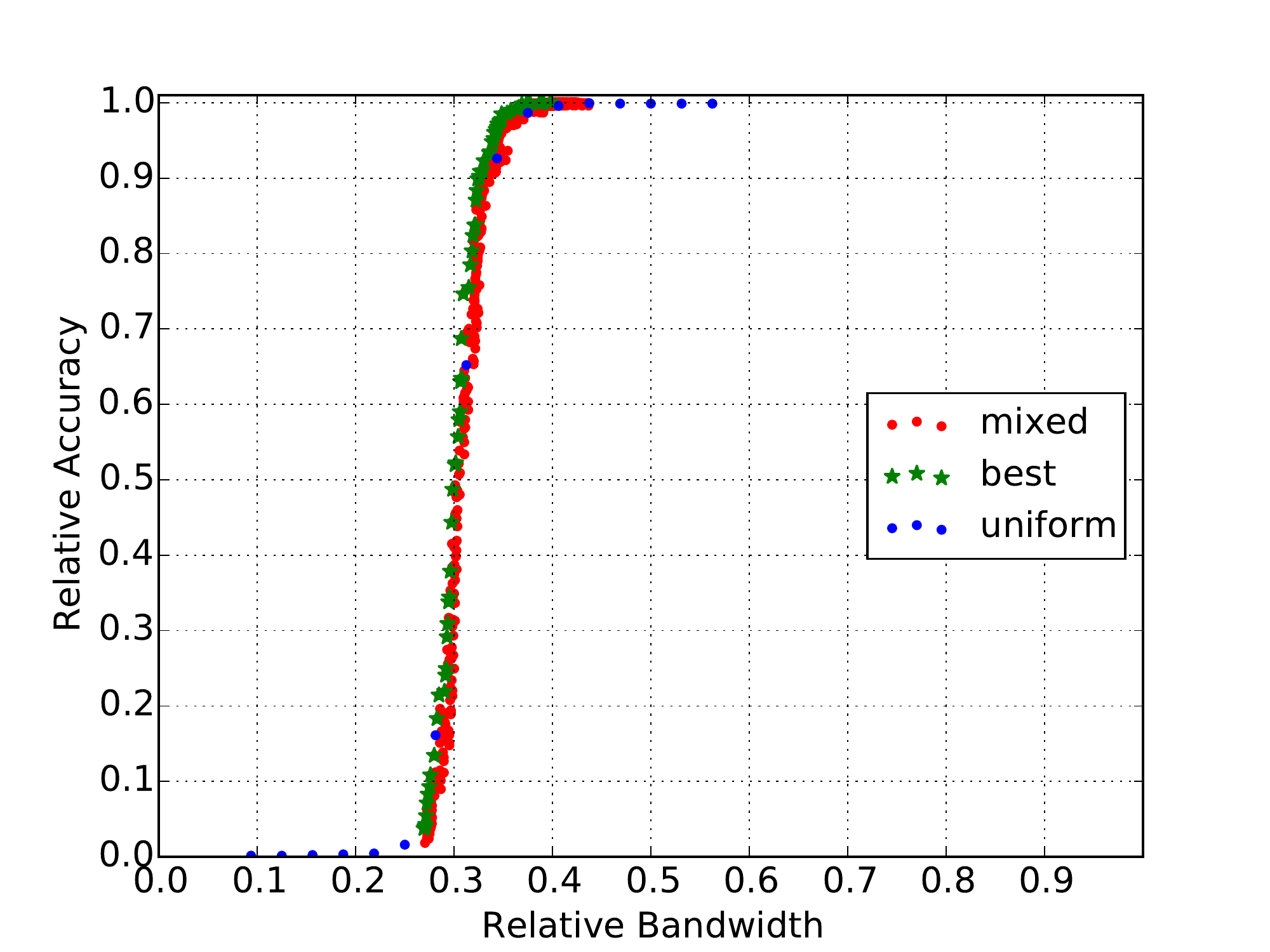}}\\
\vspace{-0.3cm}\caption{GoogLeNet}
\end{subfigure}
\caption{Design Space Exploration: Change in traffic and accuracy when using a different fixed-point representation per layer. X-axis: lower is better, Y-axis: higher is better.}
\label{fig:data-accuracy-opt}
\end{figure}

Figure \ref{fig:data-accuracy-opt} shows the results of this exploration reporting the resulting traffic (x-axis) and accuracy (y-axis) for several configurations studied.
Traffic is calculated as the number of accesses times the number of bits per element (weight or data). 

Traffic is normalized to the baseline of 32 bits per element.
Configurations are assigned to three categories: 
(1)~\textit{uniform} where all layers use the same numerical representation, 
(2)~\textit{mixed} where layers use different numerical representations, and 
(3)~\textit{best} which highlight the Pareto frontier of the mixed configurations.

Generally, the best mixed configurations achieve lower bandwidth than uniform configurations for the same accuracy.
However, there is one uniform configuration in Figure~\ref{fig:data-accuracy-opt}(d) that lies outside the Pareto frontier of the mixed configurations.
This shows that our iterative algorithm is not optimal, otherwise it would have found this better uniform configuration.
Thus, there are potentially better mixed configurations that were not explored.

Table~\ref{tab:best_mixed} reports the configurations that offer the minimum bandwidth for mixed networks given a limit on the error relative to the baseline accuracy of the network.
We expect than when accuracy can be traded for traffic savings there will be a hard constraint on the error that is acceptable.
We use 1\%-10\% as a reasonable range of tolerances that we expect for most use cases.
Beyond 10\% error the plots in Figure \ref{fig:data-accuracy-opt} tend to drop off sharply, so there is little traffic reduction for a large increase in error.
On average the optimal mixed configurations reduce the traffic by 74\% with 1\% error tolerance and 76\% over the 1\%-10\% tolerance range.
Compared to a 16-bit fixed-point baseline, the traffic reductions would be half of those reported here and thus still significant.

\begin{table}[!htbp]
    \centering
    \resizebox{0.96\textwidth}{!}{%
    \begin{tabular}{|l|l|r||l|l|r|}
       \hline
     \textbf{Tolerance} & \textbf{Bits per layer in I.F} & \textbf{TR} & \textbf{Tolerance} & \textbf{Bits per layer (I+F)} & \textbf{TR} \\ \hline
        \hline   
        \multicolumn{3}{|l||}{\textbf{LeNet}} &     \multicolumn{3}{|l|}{\textbf{AlexNet} (F=0)} \\ \hline
        1\%  &  1.1-3.1-3.0-3.0 & 0.08  &
                                                    1\% &  10-8-8-8-8-8-6-4 & 0.28 \\ \hline
        2\% &  1.1-2.0-3.0-2.0 & 0.06   &
                                                    2\% &  10-8-8-8-8-8-5-4 & 0.28 \\ \hline
        5\%  &  1.1-1.0-2.0-2.0 & 0.05  &
                                                    5\% &  10-8-8-8-7-7-5-3 & 0.28 \\ \hline
        10\%   &  1.0-2.1-3.0-2.0 & 0.05&
                                                    10\% &  9-8-8-8-7-7-5-3  & 0.26 \\ \hline
    \multicolumn{3}{|l||}{\textbf{Convnet}} & \multicolumn{3}{|l|}{\textbf{NiN} (F=0)} \\ \hline
        1\%  &  8.0-7.0-7.0-5.0-5.0 & 0.24 &
                                            1\% &   10-10-9-12-12-11-11-11-10-10-10-9  &  0.32\\ \hline
        2\%  &  7.0-8.0-6.0-4.0-4.0 & 0.22 &
                                            2\% &  10-10-9-12-12-11-11-11-10-10-10-9 & 0.32 \\ \hline
        5\%  &  7.0-8.0-5.0-3.0-3.0 & 0.22 &
                                            5\% &  10-10-10-11-10-11-11-11-10-9-9-8 & 0.32 \\ \hline
        10\%  &  7.0-8.0-5.0-3.0-3.0 & 0.22&
                                            10\%&  9-10-9-11-11-10-10-10-9-9-9-8 & 0.30 \\ \hline
	    \multicolumn{3}{|l||}{} &               \multicolumn{3}{|l|}{\textbf{GoogLeNet} (F=2)} \\ \cline{4-6}
        \multicolumn{3}{|l||}{} &                                 1\%&  14-10-12-12-12-12-11-11-11-10-9 & 0.36 \\ \cline{4-6}
        \multicolumn{3}{|l||}{} &                                2\% &  13-11-11-10-12-11-11-11-11-10-9 & 0.35 \\ \cline{4-6}
        \multicolumn{3}{|l||}{} &                               5\% &  12-11-11-11-11-11-10-10-10-9-9 & 0.34 \\ \cline{4-6}
        \multicolumn{3}{|l||}{} &                               10\% &  12-9-11-11-11-10-10-10-10-9-9 & 0.32 \\  \hline        
\end{tabular}
}
\caption{
Minimum bandwidth for mixed precision for error tolerance between  1\% and 10\%.
TR reports the traffic ratio over the 32-bit baseline.
LeNet and Convnet report the integer bits and fractional bits as $I.F$.
Fractional bits are fixed for AlexNet, NiN and GoogLeNet and the total bit width is reported.
}
\label{tab:best_mixed}
\end{table}

\section{Related Work}
\label{sec:relatedwork}
Reduced precision neural networks has been an active 
topic of research for many years \citep{Xie91trainingalgorithms,324283, 155324, And_anew,Larkin_towardshardware,Asanovic93usingsimulations,Holt93finiteprecision}.
\citet{GuptaAGN15} trains neural networks with 16-bit fixed-point numbers and stochastic rounding.
They also propose how to add hardware support for stochastic rounding.

\citet{CourbariauxBD14} used three different data formats for intermediate data: floating point, fixed point, and dynamic fixed point.
They demonstrate how networks can be trained with a low-precision data format.
However, they used a uniform representation for across the whole network.
\citet{nnfm,binaryconnect} shows that networks can be trained with binary weights without loss of accuracy. 
For MNIST, they use a fully connected network with more weights than LeNet. 
Interestingly, the total number of weight bits is comparable: 2.9 million for their network vs 3 million for LeNet with 7 bit weights. 
\citet{micro1bit} were able to reduce the precision of gradients to one bit in the training of a neural network using Stochastic Gradient Descent with almost no accuracy loss.

\citet{kyukeon_fixed_2014} and \citet{anwar_fixed_2015} quantize the signal (data) and weights in a fully connected network and CNNs and consider different quantization steps per layer. 
In the latter work, they also analyze the per-layer sensitivity to the number of quantization levels in LeNet but select the number of bits per layer manually.

\section{Conclusion}
\label{sec:conclusion}
Classification applications and quality in deep neural networks is currently limited by compute performance and the ability to communicate and store numerical data.
An effective technique for improving performance and data traffic is via the use of reduced length numerical representations.
This work provides a detailed characterization of the per-layer reduced precision tolerance of a wide range of neural networks. 
We highlight a trend of higher precision needs for newer, more complex networks. 

We proposed a method for determining an assignment of representations to layers that offers a good trade off between accuracy and data traffic.
We estimate that on average we can reduce the storage requirements for the intermediate data in our set of Convolutional Neural Networks by 74\% while maintaining classification accuracy to within 1\%.

We did not consider the effects of reduced precision during training.
The results of this work serve as motivation for studying these further. 
Promisingly, training with reduced precision has been shown to increase the network's tolerance to the error from reduced precision \citep{arc}, however, care must be taken to ensure convergence. 
The results of this work also motivate further work in exploiting the reduced precision for reducing memory bandwidth and footprint, communication bandwidth, and potentially computation bandwidth. 
The potential benefits include energy reduction, higher performance and the possibility of supporting larger networks.

\newpage

\bibliographystyle{iclr2016_conference}
\bibliography{ref}

\newpage
\appendix
\section{Supplementary Material}
\label{sec:appendix}

Table \ref{table:layers} shows the Caffe models used for each network and the Caffe layers (computaional stages) assigned to each layer in our analysis.

\begin{table*}[ht]
\centering
\begin{tabular}{|l|l|l|l|}
\hline
\textbf{Network} & \textbf{Source} & \textbf{Layer} & \textbf{Caffe Layers} \\
\hline
\hline
alexnet & http://git.io/v480W & Layer 1 & conv1,relu1,pool1,norm1 \\
\hline
 & & Layer 2 & conv2,relu2,pool2,norm2 \\
\hline
 & & Layer 3 & conv3,relu3 \\
\hline
 & & Layer 4 & conv4,relu4 \\
\hline
 & & Layer 5 & conv5,relu5,pool5 \\
\hline
 & & Layer 6 & fc6,relu6,drop6 \\
\hline
 & & Layer 7 & fc7,relu7,drop7 \\
\hline
 & & Layer 8 & fc8 \\
\hline
\hline
convnet & http://git.io/v48RM & Layer 1 & conv1,pool1,relu1 \\
\hline
 & & Layer 2 & conv2,relu2,pool2 \\
\hline
 & & Layer 3 & conv3,relu3,pool3 \\
\hline
 & & Layer 4 & ip1 \\
\hline
 & & Layer 5 & ip2 \\
\hline
\hline
googlenet & http://git.io/v480Q & Layer 1 & conv1/* \\
\hline
 & & Layer 2 & conv2/* \\
\hline
 & & Layer 3 & inception\_3a/* \\
\hline
 & & Layer 4 & inception\_3b/* \\
\hline
 & & Layer 5 & inception\_4a/* \\
\hline
 & & Layer 6 & inception\_4b/* \\
\hline
 & & Layer 7 & inception\_4c/* \\
\hline
 & & Layer 8 & inception\_4d/* \\
\hline
 & & Layer 9 & inception\_4e/* \\
\hline
 & & Layer 10 & inception\_5a/* \\
\hline
 & & Layer 11 & inception\_5b/* \\
\hline
\hline
lenet & http://git.io/v48Eu & Layer 1 & conv1,pool1 \\
\hline
 & & Layer 2 & conv2,pool2 \\
\hline
 & & Layer 3 & ip1,relu1 \\
\hline
 & & Layer 4 & ip2 \\
\hline
\hline
nin & http://git.io/v48EA & Layer 1 & conv1,relu0 \\
\hline
 & & Layer 2 & cccp1,relu1 \\
\hline
 & & Layer 3 & cccp2,relu2,pool0 \\
\hline
 & & Layer 4 & conv2,relu3 \\
\hline
 & & Layer 5 & cccp3,relu5 \\
\hline
 & & Layer 6 & cccp4,relu6,pool2 \\
\hline
 & & Layer 7 & conv3,relu7 \\
\hline
 & & Layer 8 & cccp5,relu8 \\
\hline
 & & Layer 9 & cccp6,relu9,pool3,drop \\
\hline
 & & Layer 10 & conv4-1024,relu10 \\
\hline
 & & Layer 11 & cccp7-1024,relu11 \\
\hline
 & & Layer 12 & cccp8-1024,relu12,pool4 \\
\hline
\end{tabular}
\caption{
Networks: links to sources and layer definitions.
}
\label{table:layers}
\end{table*}

\end{document}